\documentclass[11pt,a4paper]{article}
\usepackage[hyperref]{acl2020}
\usepackage{times}
\usepackage{comment}
\usepackage{latexsym}
\usepackage{amsmath,amssymb}

\usepackage{microtype}
\usepackage{amsmath}
\usepackage{todonotes}
\aclfinalcopy 

\title{Neural Language Generation: Formulation, Methods, and Evaluation}

\author{Cristina G\^arbacea$^{1}$, Qiaozhu Mei$^{1,2}$ \\
$^{1}$Department of EECS, University of Michigan, Ann Arbor, MI, USA \\
$^{2}$School of Information, University of Michigan, Ann Arbor, MI, USA \\
 {\sf \{garbacea, qmei\}@umich.edu} \\
}

\date{}

\begin{document}
\maketitle

\begin{abstract}

 
  Recent advances in neural network-based generative modeling have reignited the hopes in having computer systems capable of seamlessly conversing with humans and able to understand natural language. 
  Neural architectures have been employed to generate text excerpts to various degrees of success, in a multitude of contexts and tasks that fulfil various user needs. Notably, high capacity deep learning models trained on large scale datasets demonstrate unparalleled abilities to learn patterns in the data even in the lack of explicit supervision signals, opening up a plethora of new possibilities regarding producing realistic and coherent texts. While the field of natural language generation is evolving rapidly, there are still many open challenges to address. In this survey we formally define and categorize the problem of natural language generation. We review particular application tasks that are instantiations of these general formulations, in which generating natural language is of practical importance. Next we include a comprehensive outline of methods and neural architectures employed for generating diverse texts. Nevertheless, there is no standard way to assess the quality of text produced by these generative models, which constitutes a serious  bottleneck towards the progress of the field. To this end, we also review current approaches to evaluating natural language generation systems. We hope this survey will provide an informative overview of formulations, methods, and assessments of neural natural language generation. 
  
  

\end{abstract}

\section{Introduction}

Recent successes in deep generative modeling and representation learning have led to significant advances in natural language generation (NLG), motivated by an increasing need to understand and derive meaning from language. The research field of text generation is fundamental in natural language processing and aims to produce realistic and plausible textual content that is indistinguishable from human-written text \cite{turing1950computing}. Broadly speaking, the goal of predicting a syntactically and semantically correct sequence of consecutive words given
some context is achieved in two steps by first estimating a distribution over sentences from
a given corpus, and then sampling novel and realistic-looking sentences from the learnt distribution. Ideally, the generated sentences preserve the semantic and syntactic properties
of real-world sentences, and are different
from the training examples used to estimate the model \cite{zhang2017adversarial}. Language generation is an inherently complex task, which requires considerable linguistic and domain knowledge at multiple levels, including syntax, semantics, morphology, phonology, pragmatics, etc. Moreover, texts are generated to fulfill a communicative goal \cite{reiter2019natural}, such as to provide support in decision making, summarize content, translate between languages, converse with humans, make specific texts more accessible, as well as to entertain users or encourage them to change their behaviour. Therefore generated texts should be tailored to their specific audience in terms of appropriateness of content and terminology used \cite{paris2015user}, as well as for fairness and transparency reasons \cite{mayfield2019equity}. For a long time natural language generation models have been rule-based or relied on training shallow models on sparse high dimensional features. With the recent resurgence of neural networks, neural networks based models for text generation trained with dense vector representations have established unmatched prior performance and reignited the hopes in having machines able to understand language and seamlessly converse with humans. Indeed, generating meaningful and coherent texts is pivotal to many natural language processing tasks.   Nevertheless, designing neural networks that can generate coherent text and model long-term dependencies has long been a challenge for natural language generation due to the discrete nature of text data. Beyond that, the ability of neural network models to understand language and ground textual concepts beyond picking up on shallow patterns in the data still remains limited. Finally, evaluation of generative models for natural language is an equally active and challenging research area of significant importance in driving forward the progress of the field. 

In this work we formally define the problem of neural text generation at particular contexts and present the diverse practical applications of text generation in Section \ref{problem_definition}. In Section \ref{models} we include a comprehensive overview of deep learning methodologies and neural model architectures employed in the literature for neural network-based natural language generation. We review methods for the evaluation of the generated texts in Section \ref{evaluation}. Finally, in Section \ref{conclusion} we conclude with insights and future perspectives regarding neural text generation and evaluation. Given the rapid evolution of this field of research, we hope the current survery will serve as a thorough overview of present times neural-network based natural language generation and evaluation for anyone interested in learning about these topics, and will provide the reader with the up-to-date information on the latest research advances. Compared to the survey of \cite{gatt2018survey}, our overview is a more comprehensive and updated coverage of neural network methods and evaluation centered around the novel problem definitions and task formulations.

\section{Problem Definitions}
\label{problem_definition}
\label{problem_definition}

In what follows we formally define the natural language generation problem according to context, conditions and constraints considered when producing new text. We divide text generation into the following three categories: \textit{i)} generic or free-text generation presented in Section \ref{generic_generation}, \textit{ii)} conditional text generation introduced in Section \ref{conditional_generation}, and \textit{iii)} constrained text generation outlined in Section \ref{constrained_generation}. For each category we define the text generation problem according to the assumptions made, and clarify differences between these categories. In addition, in Section \ref{nlg_tasks} we provide examples of application areas where language generation presents rich practical opportunities.

\subsection{Generic / Free-Text Generation}
\label{generic_generation}



The problem of generic text generation aims to produce realistic text without placing any external user-defined constraints on the model output. Nevertheless, it does consider the intrinsic history of past words generated by the model as context. We formally define the problem of free-text generation. 

Given a discrete sequence of text tokens $x=(x_1, x_2, \ldots, x_n)$ as input where each $x_i$ is drawn from a fixed set of symbols, the goal of language modeling is to learn the unconditional probability distribution $p(x)$ of the sequence $x$. This distribution can be factorized using the chain rule of probability \cite{bengio2003neural} into a product of conditional probabilities:

\begin{equation}
    p(x) = \prod_{i=1}^{n} p(x_i | x_{<i})
\end{equation}
When $p(x)$ is modeled by a neural network with parameters $\theta$, the neural network is trained to minimize the negative log-likelihood over a collection of samples $D=\{x^1, \ldots, x^{|D|}\}$:

\begin{equation}
    \mathcal{L}(D)=-\sum_{k=1}^{|D|}\log p_{\theta}(x_i^k|x_{<i}^{k})
\end{equation}
Large scale models for generic text generation show promising abilities to produce coherent texts. Nevertheless, the problem of free-text generation is challenging as it places a lot of burden on the generative model  to capture complex semantic and  structural features underlying the data distribution. This can often result in incoherent and largely randomized generated text. In addition, the generated content  is uncontrollable with respect to particular attributes and modes of data being generated.

\subsection{Conditional Text Generation} 
\label{conditional_generation}




Conditional text generation is useful when generating textual content whose attributes can be controlled/ adjusted so as to enable the manipulation of the generated content. By conditioning the generative model on additional information, it becomes possible to direct the data generation process and control which modes of the data are generated. In the literature  conditional text generation is sometimes referred to as context-dependent text generation. We are aware that the word context may carry different semantics for different readers, therefore we want to clarify that in this survey the definition of conditional text generation considers as context only external attributes to the model and not any model intrinsic attributes such as for example, the history of past generated words which is already included in the formulation of the generic text generation problem in Section \ref{generic_generation}. 

Conditional language models are used to learn the distribution $p(x|c)$ of the data $x$ conditioned on a specific attribute code $c$. Similar to the formulation of generic text generation, the distribution can still be decomposed using the chain rule of probability as follows: 

\begin{equation}
\begin{split}
p(x|c)&=\prod_{i=1}^{n}p(x_i|x_{<i}, c)
\end{split}
\end{equation}
The conditional language model parameterized by a neural network can be trained with a negative log-likelihood loss function which takes into account the control code $c$:

\begin{equation}
\begin{split}
 \mathcal{L}(D)&=-\sum_{k=1}^{|D|}\log p_{\theta}(x_i^k|x_{<i}^{k}, c^{k})
\end{split}
\end{equation}

As specified above, conditional models for text generation add a contextual variable or condition into the probabilistic model transforming it into a conditional probability model. Nevertheless, conditional models do not place any hard constraints on the generated output.  
Conditioning the generated text on specific low-level attributes is approached in two ways in the literature, by either using
\textit{i)} conditional training \cite{fan2018controllable} methods which condition the model on additional control features at training time modeling $P(y|x,z)$ as a function of the output $y$ given the input $x$ and discrete control variable $z$, or via \textit{ii)} weighed decoding \cite{ghazvininejad2017hafez} methods which add control features to the decoding scoring function at test time only. Examples of attributes used for conditioning the generated text are the \textit{source sentence} in machine translation, the \textit{conversational history} in dialogue systems, the \textit{input document} in text summarization and text simplification, the \textit{input question} in question answering systems, or contextual information such as \textit{product, time and location} in review generation.  


\subsection{Constrained Text Generation}
\label{constrained_generation}

The problem of constrained text generation is focusing on generating coherent and logical texts that cover a specific set of concepts (such as pre-defined nouns, verbs, entities, phrases or sentence fragments) desired to be present
in the output, and/ or abide to user-defined rules which reflect the particular interests of the system user. Lexically constrained text generation \cite{hokamp2017lexically} places explicit constraints on independent attribute controls and combines these with differentiable approximation to produce discrete text samples. In the literature the distinction between conditional, controlled and constrained text generation is not clearly defined, and these terms are often used interchangeably. In fact, the first work that proposed generating constrained text is actually referring to the task as controlled generation \cite{hu2017toward}. In what follows we formally define the problem of constrained text generation. 

Let us consider we are (optionally) given an unordered or ordered set of $n$ concepts $x=\{c_1, c_2, \ldots, c_n\} \in \mathcal{X}$,  where $\mathcal{X}$ denotes the space of all concepts, and each $c_i \in C $, where $C$ represents the concept vocabulary and $c_i$ denotes a noun or a verb. In addition, let us assume we are also (optionally) given a set of $m$ rules $y=\{y_1, y_2, \ldots, y_m\} \in \mathcal{Y}$, with $y_i \in \mathcal{R}$, where $\mathcal{R}$ denotes the space of all rules, and each $y_i$ is a text generation constraint expressed in logical form. We formulate constrained text generation as learning the structured predictive function $f : \mathcal{X} \cup \mathcal{Y} \rightarrow \mathcal{Z}$, $\mathcal{X} \cup \mathcal{Y} \ne \phi$ which maps a set of concepts and/ or constraint rules to a generated sentence. Therefore, constrained text generation methods impose constraints on the generated sentences and produce output in the form of grammatical sentence $z \in \mathcal{Z}$ which contains all concepts present in $x$ and/ or meets all constraints specified in $y$.  The matching function $f$ manipulates the probability distribution and indicates to which extent the constraints are satisfied. In the literature, constraint text generation methods are categorized into:

\begin{itemize}
    \item  \textit{Soft-constrained text generation (priming)}: requires  the generated sentences to be semantically related to the given constraints, without strictly enforcing the presence of those constraints (for eg., topic words) in the generated content. The matching function $f$ is in this case a soft measure of semantic similarity. Typically, a corpus of (keyword, text) pairs is first constructed, followed by training a conditional text generation model to capture their co-occurence and generate text which contains the constrained keywords. Nevertheless, this approach does not guarantee that all desired keywords will be preserved during generation; some of them may get lost and will not be found in the generated output, in particular when there are constraints on simultaneously including multiple keywords. 
    
    \item  \textit{Hard-constrained text generation}: refers to the mandatory inclusion of certain keywords in the output sentences. The matching function $f$ is in this case a binary indicator, which rules out the possibility of generating infeasible sentences that do not meet the given constraints. Therefore, by placing hard constraints on the generated output,  all lexical constraints must be present in the generated output. Unlike soft-constrained models which are straightforward to design, the problem of hard-constrained text generation requires the design of complex dedicated neural network architectures. 
\end{itemize}

Constrained text generation is useful in many scenarios, such as incorporating in-domain terminology in machine translation \cite{post2018fast}, avoiding generic and meaningless responses in dialogue systems \cite{mou2016sequence}, incorporating ground-truth text fragments (such as semantic attributes, object annotations) in image caption generation \cite{anderson2017guided}. 
Typical attributes used to generate constrained natural language are the \textit{tense} and the \textit{length} of the summaries in text summarization \cite{fan2018controllable}, the \textit{sentiment} of the generated content in review generation \cite{mueller2017sequence}, \textit{language complexity} in text simplification or the \textit{style} in text style transfer applications. In addition, constrained text generation is used to overcome limitations of neural text generation models for dialogue such as genericness and repetitiveness of responses \cite{see2019makes}, \cite{serban2016building}. 

Nevertheless, generating text under specific lexical constraints
is challenging \cite{zhang2020pointer}. While for humans it is straightforward to generate sentences that cover a given set of concepts or abide to pre-defined rules by making use of their commonsense reasoning ability, generative commonsense reasoning with a constrained text generation task is not as simple for machine learning models \cite{lin2019commongen}.

\subsection{Natural Language Generation Tasks}
\label{nlg_tasks}

In what follows we present natural language generation tasks which are instances of generic, conditional and constrained text generation.  
All these applications
demonstrate the practical value of generating coherent and meaningful texts, and that advances in natural language generation are of immediate applicability and practical importance in many downstream tasks.

\subsubsection{Neural Machine Translation}
The field of machine translation is focusing on the automatic translation of textual content from one language into another language. The field has undergone major changes in recent years, with end-to-end learning approaches for automated translation based on neural networks replacing conventional phrase-based statistical methods \cite{bahdanau2014neural}, \cite{wu2016google}. In contrast to statistical models which consist of several sub-components trained and tuned separately, neural machine translation models build and train a single, large neural network end-to-end by feeding it as input textual content in the source language and retrieving its corresponding translation in the target language. Neural machine translation is a typical example of \textit{conditional text generation}, where the condition encapsulated by the conditional attribute code $c$ is represented by the input sentence in the source language and the goal task is to generate its corresponding translation in the target language. In addition, neural machine translation is also an instance of \textit{constrained text generation} given that it imposes the constraint to generate text in the target language. Additional  constraints can be placed on the inclusion in the target sentence of named entities already present in the source sentence. In what follows we formally define the problem of neural machine translation. 

We denote with $V_s$ the vocabulary of the source language and with $V_t$ the vocabulary of the target language, with $|V_t| 	\approx |V_s|$ and $V_t \cap V_s = \phi$. Let us also denote with with $V_s^*$ and $V_t^*$ all possible sentences under $V_s$, respectively $V_t$. Given a source sentence $X=(x_1, x_2, \ldots, x_l), X \in V_s^*, x_i \in V_s$, where $x_i$ is the $i^{th}$ word in $X$, $ \forall i=1,\ldots, l$, the goal is to generate the distribution over the possible output sentences $Y=(y_1, y_2, \ldots, y_{l^{'}}), Y \in V_t^*, y_j \in V_t$,  where $y_j$ is the $j^{th}$ word in $Y$, $ \forall j=1,\ldots, l^{'}$ by factoring $Y$ into a chain of conditional probabilities with left-to-right causal structure using a neural network with parameters $\theta$:

\begin{equation}
    p(Y|X; \theta) = \prod_{t=1}^{l^{'} + 1} p(y_t | y_{0:t-1}, x_{1:l}; \theta)
\end{equation}
Special sentence delimiters $y_0(<$S$>)$ and $y_{l^{'} + 1}(<$E$>)$ are commonly added to the vocabulary to mark the beginning and end of target sentence $Y$. Typically in machine translation the source and target vocabularies consist of the most frequent words used in a language (for eg., top 15,000 words), while the remaining words are replaced with a special $<$UNK$>$ token. Every source sentence $X$ is usually mapped to exactly one target sentence $Y$, and there is no sharing of words between the source sentence $X$ and the target sentence $Y$.

Although neural network based approaches to machine translation have resulted in superior performance compared to statistical models, they are computationally expensive both in training and in translation inference time. The output of machine translation models is evaluated by asking human annotators to rate the generated translations on various dimensions of textual quality, or by comparisons with human-written reference texts using automated evaluation metrics.

\subsubsection{Text Summarization}
\label{text_summarization}

Text summarization is designed to facilitate a quick grasp of the essence of an input document by producing a condensed summary of its content. This can be achieved in two ways, either by means of extractive summarization or through abstractive/generative summarization. While extractive summarization \cite{nallapati2017summarunner} methods produce summaries by copy-pasting the relevant portions from the input document, abstractive summarization \cite{rush2015neural}, \cite{nallapati2016abstractive}, \cite{see2017get} algorithms can generate novel content that is not present in the input document. Hybrid approaches combining extractive summarization techniques with a a neural abstractive summary generation serve to identify salient information in a document and generate distilled Wikipedia articles \cite{liu2018generating}. Characteristics of a good summary include brevity, fluency, non-redundancy, coverage and logical entailment of the most salient pieces of information from the input document(s). 

Text summarization is a \textit{conditional text generation task} where the condition is represented by the given document(s) to be summarized. Additional control codes are used in remainder summarization  offering flexibility to define which parts of the document(s) are of interest, for eg., remaining paragraphs the user has not read yet, or in source-specific summarization to condition summaries on the source type of input documents, for eg., newspapers, books or news articles.
Besides being a conditional text generation task, text summarization is also a typical example of \textit{constrained text generation} where the condition is set such that the length of the resulting summary is strictly less than the length of the original document. Unlike machine translation where output length varies depending on the source content, in text summarization the length of the output is fixed and pre-determined. 
Controlling the length of the generated summary allows to digest information at different levels of granularity and define the level of detail desired accounting for particular user needs and time budgets; for eg., a document can be summarized into a headline, a single sentence or a multi-sentence paragraph. In addition, explicit constraints can be placed on specific concepts desired for inclusion in the summary. Most frequently, named entities are used as constraints in text summarization to ensure the generated summary is specifically focused on topics and events describing them. In addition, in the particular case of extractive summarization, there is the additional constraint that sentences need to be picked explicitly from the original document. In what follows we formally define the task of text summarization.



We consider the input consisting of a sequence of $M$ words $x=(x_1, x_2, \ldots, x_M), x_i \in \mathcal{V_\mathcal{X}}, i=1, \ldots, M$, where $\mathcal{V_\mathcal{X}}$ is a fixed vocabulary of size $|\mathcal{V}_\mathcal{X}|$. Each word $x_i$ is represented as an indicator vector $x_i \in \{0,1\}^{\mathcal{V}_\mathcal{X}}$, sentences are represented as sequences of indicators and $\mathcal{X}$ denotes the set of all possible inputs. A summarization model takes $x$ as input and yields a shorter version of it in the form of output sequence $y=(y_1, y_2, \ldots, y_N)$, with $N<M$ and $y_j \in \{0,1\}^{\mathcal{V_\mathcal{Y}}}, \forall j=1, \ldots, N$. 



\textit{Abstractive / Generative Summarization} We define $\mathcal{Y} \subset (\{0, 1\}^{\mathcal{V}_\mathcal{Y}}, \ldots, \{0, 1\}^{\mathcal{V}_{\mathcal{Y}}})$ as the set of all possible generated summaries of length $N$, with $y \in \mathcal{Y}$. The summarization system is abstractive if it tries to find the optimal sequence $y^*, y^* \subset \mathcal{Y}$, under the scoring function $s : \mathcal{X} \times \mathcal{Y} \rightarrow \mathcal{R}$, which can be expressed as: 

\begin{equation}
    y^* = \arg \max_{y \in \mathcal{Y}} s(x, y)
\end{equation}

\textit{Extractive Summarization} As opposed to abstractive approaches which generate novel sentences, extractive approaches transfer parts from the input document $x$ to the output $y$:

\begin{equation}
    y^* = \underset{m \in \{1, \ldots, M\}^{N}}{\arg\max} s(x, x_{[m_1, \dots, m_N]})
\end{equation}
Abstractive summarization is notably more challenging than extractive summarization, and allows to incorporate real-world knowledge, paraphrasing and generalization, all crucial components of high-quality summaries \cite{see2017get}. In addition, abstractive summarization
does not impose any hard constraints on the system output other than shorter length and gives the system a lot of freedom to generate suitable content, which in turn results in the system's ability to fit a wide range of training data. Approaches for neural abstractive summarization build upon advances in machine translation. Attention mechanisms \cite{bahdanau2014neural} and pointer networks \cite{vinyals2015pointer} are used to focus on specific parts of the source document and copy input entities. Nevertheless, a common limitation of current abstractive summarization models is their tendency to copy  long passages from
the source document as opposed to generating novel content. Consequently, the word overlap between the source document and the generated abstractive summary is generally high \cite{see2017get}, \cite{kryscinski2018improving}.

Very related in nature to the task of text summarization is the problem of \textit{text compression}, which takes as input a text document and aims to produce a short summary of it by deleting the least critical information, while retaining the most important ideas and preserving sentence fluency. Sentence compression is referred in the literature as a ``scaled down version of the text summarization problem''\cite{knight2002summarization}, and is of practical importance in many downstream applications including text summarization, subtitle generation and displaying text on small screens \cite{mallinson2018sentence}. 
Similar to text summarization, text compression is both a \textit{conditional text generation} and \textit{constrained text generation} task. The condition is represented by the input document for which the text compression system needs to output a condensed version. The task is also \textit{constrained text generation} given the system needs to produce a compressed version of the input strictly shorter lengthwise. In addition, there can be further constraints specified when the text compression output is desired to be entity-centric.  

We denote with $C_i=\{c_{i1}, c_{i2}, \ldots, c_{il}\}$ the set of possible compression spans and with $y_{y,c}$ a binary variable which equals 1 if the $c^{th}$ token of the $i^{th}$ sentence $\hat{s_i}$ in document $D$ is deleted, we are interested in modeling the probability $p(y_{i,c}|D, \hat{s_i})$. Following the same definitions from section \ref{text_summarization}, we can formally define the optimal compressed text sequence under scoring function $s$ as:

\begin{equation}
    y^* =\underset{m \in \{1, \ldots, M\}^{N}, m_{i-1} < m_i}{\arg \max} s(x, x_{[m_1, \dots, m_N]})
\end{equation}


Shorter paraphrases are generated through end-to-end neural networks in \cite{filippova2015sentence}. Unsupervised models based on denoising autoencoders wihout the need for paired corpora are used in \cite{fevry2018unsupervised}. The level of compression is computed as the character length ratio between the source sentence and the target sentence \cite{martin2019controllable}. Sentence compressions  are identified by constituency parsing and scored by neural models for text summarization \cite{xu2019neural}. Factors directly impacting the generated summary complexity  where are the compression rate, the summarization technique and the nature of the summarized corpus \cite{vodolazova2019towards}. 

Current datasets, models and evaluation metrics for text summarization are considered not robust enough \cite{kryscinski2019neural}. Shortcomings include uncurated automatically collected datasets, models that overfit to biases in the data and produce outputs with little diversity, as well as  non-informative evaluation metrics weakly correlated with human judgements.

\subsubsection{Text Simplification}

Text simplification is designed to reduce the lexical and syntactic complexity of text, while preserving the main idea and approximating the original meaning. The goal of text simplification systems is to make highly specialized textual content accessible to readers who lack adequate literacy skills, such as children, people with low
education, people who have reading disorders or dyslexia, and
non-native speakers of the language. In the literature text simplification has been addressed at multiple levels: \textit{i)} lexical simplification \cite{devlin1999simplifying} is concerned with replacing complex words
or phrases with simpler alternatives; \textit{ii)} syntactic simplification \cite{siddharthan2006syntactic}
alters the syntactic structure of the sentence; \textit{iii)} semantic simplification \cite{kandula2010semantic}, sometimes also known as explanation generation, paraphrases
portions of the text into simpler and clearer variants. More recently, end-to-end models for text simplification attempt to address all these steps at once. 

Text simplification is an instance of \textit{conditional text generation} given we are conditioning on the input text to produce a simpler and more readable version of a complex document, as well as an instance of \textit{constrained text generation} since there are constraints on generating simplified text that is shorter in length compared to the source document and with higher readability level. To this end, it is mandatory to use words of lower complexity from a much simpler target vocabulary than the source vocabulary. We formally introduce the text simplification task below.

Let us denote with $V_s$ the vocabulary of the source language and with $V_t$ the vocabulary of the target language, with $|V_t| \ll |V_s|$ and $V_t \subseteq V_s$. Let us also denote with with $V_s^*$ and $V_t^*$ all possible sentences under $V_s$, respectively $V_t$. Given source sentence $X=(x_1, x_2, \ldots, x_l), X \in V_s^*, x_i \in V_s$, where $x_i$ is the $i^{th}$ word in $X$, $ \forall i=1,\ldots, l$, the goal is to produce the simplified sentence $Y=(y_1, y_2, \ldots, y_{l^{'}}), Y \in V_t^*, y_j \in V_t$,  where $y_j$ is the $j^{th}$ word in $Y$, $ \forall j=1,\ldots, l^{'}$ by modeling the conditional probability $p(Y |X)$. In the context of neural text simplification, a neural network with parameters $\theta$ is used to maximize the probability $p(Y|X; \theta)$.    

Next we highlight differences between machine translation and text simplification. Unlike machine translation where the output sentence $Y$ does not share any common terms with the input sentence $X$, in text simplification some or all of the words in $Y$ might remain identical with the words in $X$ in cases when the terms in $X$ are already simple. In addition, unlike machine translation where the mapping between the source sentence and the target sentence is usually one-to-one, in text simplification the relation between the source sentence and the target sentence can be one-to-many or many-to-one, as simplification involves splitting and merging operations \cite{surya2018unsupervised}. Furthermore, infrequent words in the vocabulary cannot be simply dropped out and replaced with an unknown token as it is typically done in machine translation, but they need to be simplified appropriately corresponding to their level of complexity \cite{wang2016text}. Lexical simplification and content reduction is simultaneously approached with neural machine translation models in \cite{nisioi2017exploring}, \cite{sulem2018simple}. Nevertheless, text simplification presents particular challenges compared to machine translation. First, simplifications need to be adapted to particular user needs, and ideally personalized to the educational background of the target audience \cite{bingel2018personalized}, \cite{mayfield2019equity}.
Second, text simplification has the potential to bridge the communication gap between specialists and laypersons in many scenarios. For example, in the medical domain it can help improve the understandability  of clinical records \cite{shardlow2019neural}, address disabilities and inequity in educational environments \cite{mayfield2019equity}, and assist with providing accessible and timely information to the affected population in crisis management \cite{temnikova2012text}. 

\subsubsection{Text Style Transfer}

Style transfer is a newly emerging task designed to preserve the information content of a source sentence while delivering it to meet desired presentation constraints. To this end, it is important to disentangle the content itself from the style in which it is presented and be able to manipulate the style so as to easily change it from one attribute into another attribute of different or opposite polarity. This is often achieved without the need for parallel data for source and target styles, but accounting for the constraint that the transferred sentences should match in style example sentences from the target style. To this end, text style transfer is an instance of \textit{constrained text generation}. In addition, it is also a typical scenario of \textit{conditional text generation} where we are conditioning on the given source text. 
Style transfer has been originally used in computer vision applications for image-to-image translation \cite{gatys2016image}, \cite{liu2016coupled}, \cite{zhu2017unpaired}, and more recently has been used in natural natural language processing applications for machine translation, sentiment modification to change the sentiment of a sentence from positive to negative and vice versa, word substitution decipherment and word order recovery \cite{hu2017toward}. 

The problem of style transfer in language generation can be formally defined as follows. Given two datasets $X_1=\{x_1^{(1)}, x_1^{(2)}, \ldots, x_1^{(n)} \}$ and $X_2=\{x_2^{(1)}, x_2^{(2)}, \ldots, x_2^{(n)} \}$ with the same content distribution but different unknown styles $y_1$ and $y_2$, where the samples in dataset $X_1$ are drawn from the distribution $p(x_1|y_1)$ and the samples in dataset $X_2$ are drawn from the distribution $p(x_2|y_2)$, the goal is to estimate the style transfer functions between them $p(x_1|x_2; y_1, y_2)$ and $p(x_2|x_1; y_1, y_2)$. According to the formulation of the problem we can only observe the marginal distributions $p(x_1|y_1)$ and $p(x_2|y_2)$, and the goal is to recover the joint distribution $p(x_1, x_2|y_1, y_2)$, which can be expressed as follows assuming the existence of latent content variable $z$ generated from distribution $p(z)$:

\begin{equation}
    p(x_1, x_2|y_1, y_2) = \int_z p(z) p(x_1|y_1, z) p(x_2|y_2, z) dz 
\end{equation}
Given that $x_1$ and $x_2$ are independent from each other given $z$, the conditional distribution corresponding to the style transfer function is defined:

\begin{equation}
\begin{split}
    p(x_1|x_2;y_1, y_2)&= \int_z p(x_1,z|x_2; y_1,y_2)dz \\
    &=\int_z p(x_1|y_1,z) p(x_2|y_2,z) dz \\
    &=\mathbb{E}_{z \sim p(z|x_2, y_2)}[p(x_1|y_1, z)]
\end{split}
\end{equation}
Models proposed in the literature for style transfer rely on encoder-decoder models. Given encoder $E:X \times Y \rightarrow Z$ with paramters $\theta_E$ which infers the content $z$ and style $y$ for a given sentence $x$, and generator $G:Y \times Z\rightarrow X$ with parameters $\theta_G$ which given content $z$ and style $y$ generates sentence $x$, the reconstruction loss can be defined as follows:

\begin{equation}
\begin{split}
    \mathcal{L}_{\text{rec}} = &\mathbb{E}_{x_1 \sim X_1}[-\log p_G (x_1|y_1, E(x_1, y_1))] + \\
    & \mathbb{E}_{x_2 \sim X_2}[-\log p_G (x_2|y_2, E(x_2, y_2))]
\end{split}
\end{equation}
Latent VAE representations are manipulated to generate textual output with specific attributes, for eg. contemporary text written in Shakespeare style or improving the  positivity sentiment of a sentence \cite{mueller2017sequence}. Style-independent content representations are learnt via disentangled latent representations for generating sentences
with controllable style attributes \cite{shen2017style}, \cite{hu2017toward}. Language models are employed as style discriminators to learn disentangled representations for unsupervised text style transfer tasks such as sentiment modification \cite{yang2018unsupervised}.  

\subsubsection{Dialogue Systems}


A dialogue system, also known as a conversational agent, is a computer system designed to converse with humans using natural language. To be able to carry a meaningful conversation with a human user, the system needs to first understand the message of the user, represent it internally, decide how to respond to it and issue the target response  using natural language surface utterances \cite{chen2017survey}. Dialogue generation is an instance of \textit{conditional text generation} where the system response is conditioned on the previous user utterance and frequently on the overall conversational context. Dialogue generation can also be an instance of \textit{constrained text generation} when the conversation is carried on a topic which explicitly involves entities such as locations, persons, institutions, etc. 
From an application point of view, dialogue systems can be categorized into \cite{keselj2009speech}:

\begin{itemize}
  \item \textit{task-oriented dialogue agents}: are designed to have short conversations with a human user to help him/ her complete a particular task. For example, dialogue agents embedded into digital assistants and home controllers assist with
  finding products, booking accommodations, provide travel directions, make restaurant reservations and phone calls on behalf of their users. Therefore, task-oriented dialogue generation is an instance of both conditional and constrained text generation. 
  \item \textit{non-task oriented dialogue agents or chat-bots}: are designed for carrying extended conversations with their users on a wide range of open domains. They are set up to mimic human-to-human interaction and unstructured human dialogues in an entertaining way. Therefore, non-task oriented dialogue is an instance of conditional text generation. 
\end{itemize}


 We formally define the task of dialogue generation.
Generative dialogue models take as input a dialogue context $c$ and generate the next response $x$. The training data consists of a set of samples of the form $\{c^{n}, x^{n}, d^{n}\} \sim p_{\text{source}}(c,x,d)$, where $d$ denotes the source domain. At testing time, the model is given the dialog context $c$ and the target domain, and must generate the correct response $x$. The goal of a generative dialogue model is to learn the function $\mathcal{F}: C \times D \rightarrow X$ which performs well on unseen examples from the target domain after seeing the training examples on the source domain. The source domain and the target domain can be identical; when they differ the problem is defined as zero-shot dialogue generation \cite{zhao2018zero}. The dialogue generation problem can be summarized as:

\begin{equation}
\begin{split}
    \text{Training data}&: \{c^{n}, x^{n}, d^{n}\} \sim p_{\text{source}}(c,x,d) \\
    \text{Testing data}&: \{c, x, d\} \sim p_{\text{target}}(c,x,d) \\
    \text{Goal}&: \mathcal{F}: C \times D \rightarrow X
\end{split}
\end{equation}
A common limitation of neural networks for dialogue generation is that they tend to generate safe, universally relevant responses that carry little meaning \cite{serban2016building}, \cite{li2016diversity}, \cite{mou2016sequence}; for example universal replies such as ``I don't know'' or ``something'' frequently occur in the training set are likely to have high estimated probabilities at decoding time.
Additional factors that impact the conversational flow in generative models of dialogue are identified as repetitions and contradictions of previous statements, failing to balance specificity
with genericness of the output, and not taking turns in asking questions \cite{see2019makes}.  
Furthermore, it is desirable for generated dialogues to incorporate explicit personality traits \cite{zheng2019personalized} and control the sentiment \cite{kong2019adversarial} of the generated response to resemble human-to-human conversations.

\subsubsection{Question Answering}

Question answering systems are designed to find and integrate information from various sources to provide responses to user questions  \cite{fu2018natural}. While traditionally candidate answers consist of words, phrases or sentence snippets retrieved and ranked appropriately from knowledge bases and textual documents \cite{kratzwald2019rankqa}, answer generation aims to produce more natural answers by using neural models to generate the answer sentence. Question answering can be considered as both a \textit{conditional text generation} and \textit{constrained text generation} task. A question answering system needs to be conditioned on the question that was asked, while simultaneously ensuring that concepts needed to answer the question are found in the generated output.

A question answering system can be formally defined as follows. Given a context paragraph $C=\{c_1, c_2, \ldots, c_n\}$ consisting of $n$ words from word vocabulary $\mathcal{V}$ and the query $Q=\{q_1, q_2, \ldots, q_m\}$ of $m$ words in length, the goal of a question answering system is to either: \textit{i)} output a span $S=\{c_i, c_{i+1}, \ldots, c_{i+j}\}, \forall i=1, \ldots, n$ and $\forall j=0, \ldots, n-i$ from the original context paragraph $C$, or \textit{ii)} generate a sequence of words $A=\{a_1, a_2, \ldots, a_l\}, a_k \in \mathcal{V}, \forall k=1, \ldots, l$ as the output answer. Below we differentiate between multiple types of question answering tasks:

\begin{itemize}
    \item \textit{Factoid Question Answering}: given a description of an entity (person, place or item) formulated as a query and a text document, the task is to identify the entity referenced in the given piece of text. This is an instance of both conditional and constrained text generation, given conditioning on the input question and constraining the generation task to be entity-centric.  Factoid question answering methods combine word and phrase-level representations across sentences to reason about entities \cite{iyyer2014neural}, \cite{yin2015neural}.
    
    \item \textit{Reasoning-based Question Answering}: given a collection of documents and a query, the task is to 
    reason, gather, and synthesize disjoint pieces of information spread within documents and across multiple documents to generate an answer  \cite{de2019question}. The task involves multi-step reasoning and understanding of implicit relations for which humans typically rely on their background commonsense knowledge \cite{bauer2018commonsense}. The task is conditional given that the system generates an answer conditioned on the input question, and may be constrained when the information across documents is focused on entities or specific concepts that need to be incorporated in the generated answer.
    
    
    \item \textit{Visual Question Answering}: given an image and a natural language question about the image, the goal is to provide an accurate natural language answer to the question posed about the image \cite{antol2015vqa}. By its nature the task is conditional, and can be constraint when specific objects or entities in the image need to be included in the generated answer. 
\end{itemize}

Question answering systems that meet various information needs are proposed in the literature, for eg., for answering mathematical questions \cite{schubotz2018introducing}, medical information needs \cite{wiese2017neural}, \cite{bhandwaldar2018uncc}, quiz bowl questions \cite{iyyer2014neural}, cross-lingual and multi-lingual questions \cite{loginova2018towards}. In practical applications of question answering, users are typically not only interested in learning the exact answer word, but also in how this is related to other important background information and to previously asked questions and answers \cite{fu2018natural}. 

\subsubsection{Image / Video Captioning}

Image captioning is designed to generate captions in the form of textual descriptions for an image. This involves the recognition of the important objects present in the image, as well as object properties and interactions between objects to be able to generate syntactically and semantically correct natural language sentences  \cite{hossain2019comprehensive}. In the literature the image captioning task has been framed from either a natural language generation perspective \cite{kulkarni2013babytalk}, \cite{chen2017sca} where each system produces a novel sentence, or from a ranking perspective where existing captions are ranked and the top one is selected \cite{hodosh2013framing}. Image/ video captioning is a \textit{conditional text generation} task where the caption is conditioned on the input image or video. In addition, it can be a \textit{constrained text generation} task when specific concepts describing the input need to be present in the generated output.

Formally, the task of image/ video captioning takes as input an image or video $I$ and generates a sequence of words $y=(y_1, y_2, \ldots, y_N), y \in V^{*} \text{ and } y_i \in V, \forall i=1, \ldots, N$, where $V$ denotes the vocabulary of output words and includes special tokens to mark the beginning $<$S$>$ and end $<$E$>$ of a sentence, as well as the unknown token $<$UNK$>$ used for all words not present in the vocabulary $V$, and $V^{*}$ denotes all possible sentences over $V$. Given training set $\mathcal{D}=\{(I, y^{*})\}$ containing $m$ pairs of the form $(I_{j}, y_{j}^{*}), \forall j=1, \ldots, m$ consisting of input image $I_{j}$ and its corresponding ground-truth caption $y_{j}^{*}=(y_{j_{1}}^{*}, y_{j_{2}}^{*}, \ldots, y_{j_{M}}^{*}), y_{j}^{*} \in V^{*} \text{ and } y_{j_{k}}^{*} \in V, \forall k=1, \ldots, M$, we want to maximize the probabilistic model $p(y|I; \theta)$ with respect to model parameters $\theta$.

\subsubsection{Narrative Generation / Story Telling}

Neural narrative generation aims to produce coherent stories automatically and is regarded as an important step towards computational creativity \cite{gervas2009computational}. Unlike machine translation which produces a complete transduction of an input sentence which fully defines the target semantics, story telling is a long-form open-ended text generation task which simultaneously addresses two separate challenges: the selection of appropriate content (\textit{``what to say''}) and the surface realization of the generation (\textit{``how to say it''})\cite{wiseman2017challenges}. In addition, the most difficult aspect of neural story generation is producing a a coherent and fluent story which is much longer than the short input specified by the user as the story title. To this end, many neural story generation models assume the existence of a high-level plot (commonly specified as a one-sentence outline) which serves the role of a bridge between titles and stories \cite{chen2019learning}, \cite{fan2018hierarchical}, \cite{xu2018skeleton}, \cite{drissi2018hierarchical}, \cite{yao2019plan}. Therefore, narrative generation is a \textit{constrained text generation} task since  explicit constraints are placed on which concepts to include in the narrative so as to steer the generation in particular topic directions. In addition, another constraint is that the output length needs to be strictly greater than the input length. We formally define the task of narrative generation below.  

Assuming as input to the neural story generation system the title $x=x_1, x_2, \ldots, x_I$ consisting of $I$ words, the goal is to produce a comprehensible and logical story $y=y_1, y_2, \ldots, y_J$ of $J$ words in length. Assuming the existence of a one sentence outline $z=z_1, z_2, \ldots, z_K$ that
contains $K$ words for the entire story, the latent variable model for neural story generation can be formally expressed:

\begin{equation}
    P(y|x; \theta, \gamma) = \sum_z P(z|x; \theta) P(y|x,z; \gamma)
\end{equation}
where $P(z|x; \theta)$ defines a planning model parameterized by $\theta$ and $P(y|x,z; \gamma)$ defines a generation model parameterized by $\gamma$.

The planning model $P(z|x; \theta)$ receives an input the one sentence title $z$ for the narrative and generates the narrative outline given the title: 

\begin{equation}
    P(z|x; \theta) = \prod_{k=1}^{K} P(z_k|x, z_{<k}; \theta)
\end{equation}
where $z_{<k}=z_1, z_2, \ldots, z_{k-1}$ denotes a partial outline. The generation model is used to produce a narrative given a title and an outline:  

\begin{equation}
    P(y|x,z; \gamma) = \prod_{j=1}^{J}P(y_j|x,z, y_{<j}; \gamma)
\end{equation}
where $y_{<j}$ denotes a partially generated story.

Hierarchical models for story generation break down the generation process into multiple steps: first modelling the action sequence, then the
story narrative, and finally entities such as story
characters \cite{fan2019strategies}. While existing models can generate stories with
good local coherence,  generating long stories is challenging. Difficulties in coalescing individual phrases into coherent plots and in maintaining character consistency throughout the story lead to a rapid decrease in coherence as the output length increases \cite{van2019narrative}.
Neural narrative generation combining story-writing with human collaboration in an interactive way improves both story quality and human engagement \cite{goldfarb2019plan}.

\subsubsection{Poetry Generation}

Automatic poetry generation is an important step towards computational creativity. In the poem generation literature, the generator operates in an interactive context where the user initially supplies the model with a set of keywords representing the concepts which outline the main writing intents, as well as their ordering. The user is also in charge of selecting a particular format for the generated poem. For example, common formats are quatrain, consisting of 4 lines of sentences, or regulated verse in which the poem is made up of 8 lines of sentences. The process is interactive and the author can keep modifying terms to reflect his writing intent. Poetry generation is a \textit{constrained text generation} problem since user defined concepts need to be included in the generated poem. At the same time, it can also be a \textit{conditional text generation problem} given explicit conditioning on the stylistic features of the poem. We define the petry generation task below. 

Given as input a set of keywords that summarize an author's writing intent $K=\{k_1, k_2, \ldots, k_{|K|}\}, \text{ where each } k_i \in V, i=1, \ldots, |K|$ is a keyword term from vocabulary $V$, the goal is to generate a poem $\mathcal{P} = \{w|w \in \Omega \}$ where each term $w$ is selected from the candidate term set $\Omega=\{w|w \in \{K \cup \{V-K\}\}\}$, and $ K \subseteq P, P \subseteq \Omega$ to fit the user specified constraints of the poetry format. The generative model computes the probability of line $S_{i+1}=w_1, w_2, \ldots, w_m$ given all previously generated poem lines $S_{1:i}, i \ge 1$ (or alternatively, only the previous generated line or lexical n-grams) as follows:

\begin{equation}
    P(S_{i+1}|S_{1:i})= \prod_{j=1}^{m-1} P(w_{j+1} | w_{1:j}, S_{1:i})
\end{equation}

Poetry composition is formulated as a constrained optimization problem in a generative summarization framework in iPOET \cite{yan2013poet}. Candidate terms from a large human-written poem corpus are retrieved to match the user intent, and then clustered to fit the poetry format, tone, rhythm, etc. Each cluster generates one line of the poem in a multi-pass generative summarization framework by conducting iterative term substitutions so that the generated poem matches the initial user constraints and poetic preference, and the relevance and coherence of the output is maximized. Generative models that jointly perform content selection and surface realization are proposed in
\cite{zhang2014chinese}. Generated poems are revised and polished through
multiple style configurations in \cite{ghazvininejad2017hafez}. Neural poetry generation models based on maximum likelihood estimation (MLE) only learn the most common patterns in the poetry corpus and generate outputs with little diversity \cite{zhang2017flexible}. In addition, these MLE based models suffer from loss-evaluation mismatch \cite{wiseman2016sequence} manifested through incompatibility at evaluation time between the word-level loss function optimized by MLE and humans focusing on whole sequences of poem lines and assessing fine-grained criteria of the generated text such as fluency, coherence, meaningfulness and overall quality. These human evaluation criteria are modeled and incorporated into the reward function of a mutual reinforcement learning framework for poem generation \cite{yi2018automatic}. For a detailed overview of poetry generation we point the reader to \cite{oliveira2017survey}.

\subsubsection{Review Generation}

Product reviews allow users to express opinions for different aspects of products or services received, and are popular on many online review websites such as Amazon, Yelp, Ebay, etc. These online reviews encompass a wide variety of writing styles and polarity strengths. The task of review generation is similar in nature to sentiment analysis and a lot of past work has focused on identifying and extracting subjective content in review data \cite{liu2015sentiment}, \cite{zhao2016sentiment}. Automatically generating reviews given contextual information focused on product attributes, ratings, sentiment, time and location is a meaningful \textit{conditional text generation} task. Common product attributes used in the literature are the user ID, the product ID, the product rating or the user sentiment for the generated review \cite{dong2017learning}, \cite{tang2016context}. The task can also be \textit{constrained text generation} when topical and syntactic characteristics of natural
languages are explicitly specified as constraints to incorporate in the generation process. We formally define the review generation task below.

Given as input a set of product  attributes $a=(a_1, a_2, \ldots, a_{|a|})$ of fixed length $|a|$, the goal is to generate a product review $r=(y_1, y_2, \ldots, y_{|r|})$ of variable length $|r|$ by maximizing the conditional probability $p(r|a)$:

\begin{equation}
    p(r|a)= \prod_{t=1}^{|r|}p(y_t|y_{<t}, a)
\end{equation}
where $y_{<t} = (y_1, y_2, \ldots, y_{t-1})$.  The training data consists of pairs $(a, r)$ of attributes $a$ with their corresponding reviews $r$, and the model learns to maximize the likelihood of the generated reviews given the input attributes for the training data $\mathcal{D}$. The optimization problem can therefore be expressed as:

\begin{equation}
    \max \sum_{(a,r) \in \mathcal{D}} \log p(r|a) 
\end{equation}

Generating long, well-structured and informative reviews requires considerable effort when written by human users and is a similarly challenging task to do automatically  \cite{li2019generating}.





\subsubsection{Miscellaneous tasks related to natural language generation}

Handwriting synthesis aims to automatically generate data that resembles natural handwriting and is a key component in the development of intelligent systems that can provide personalized experiences to humans  \cite{zong2014strokebank}. The task of handwritten text generation is very much analogous to sequence generation. Given as input a user defined sequence of words $x=(x_1, x_2, \ldots, x_T)$ which can be either typed into the computer system or fed as an input image $I$ to capture the user's writing style, the goal of handwriting generation is to train a neural network model which can produce a cursive handwritten version of the input text to display under the form of output image $O$ \cite{graves2013generating}. Handwriting generation is a \textit{conditional generation task} when the system is conditioning on the input text. In addition, it is also a \textit{constrained text generation task} since the task is constrained on generating text in the user's own writing style. While advances in deep learning have given computers the ability to see and recognize printed text from input images, generating cursive handwriting is a considerably more challenging problem \cite{alonso2019adversarial}. Character boundaries are not always well-defined, which makes it hard to segment handwritten text into individual pieces or characters. In addition, handwriting evaluation is ambiguous and not well defined given the multitude of existent human handwriting style profiles \cite{mohammed2018handwriting}.

Other related tasks where natural language generation plays an important role are generating questions, arguments, counter-arguments and opinions, news headlines and digests, reports, financial statements, stock market reports, sports reports, slides and entire presentations, error corrections, generating creative and entertaining texts, composing music, lyrics and tweets, data-to-text generation, paraphrasing, speech synthesis, generating proteins as sequences of aminoacids, code in a programming language of choice, etc. All these tasks illustrate the widespread importance of having robust models for natural language generation.



\section{Models}
\label{models}

Neural networks are used in a wide range of supervised and unsupervised machine learning tasks due to their ability to learn hierarchical representations from raw underlying features in the data and model complex high-dimensional distributions. A wide range of model architectures based on neural networks have been proposed for the task of natural language generation in a wide variety of contexts and applications. In what follows we briefly discuss the main categories of generative models in the literature and continue with presenting specific models for neural language generation.

Deep generative models have received a lot of attention recently due to their ability to model complex high-dimensional distributions. These models combine uncertainty estimates provided by probabilistic models with the flexibility and scalability of deep neural networks to learn in an unsupervised way the distribution from which data is drawn. Generative probabilistic models are useful for two reasons: \textit{i)} can perform density estimation and
inference of latent variables, and \textit{ii)} can sample efficiently from the probability density represented by the input data and generate novel content. Deep generative models can be classified into either explicit or implicit density probabilistic models. On the one hand, \textit{explicit density models} provide an explicit parametric specification of the data distribution and have tractable likelihood functions. On the other hand, \textit{implicit density models} do not specify the underlying distribution of the data, but instead define a stochastic process which allows to simulate the data distribution after training by drawing samples from it. Since the data distribution is not explicitly specified, implicit generative models do not have a tractable likelihood function. A mix of both explicit and implicit models have been used in the literature to generate textual content in a variety of settings. Among these, we enumerate explicit density models with tractable density such as autoregressive models \cite{bahdanau2014neural}, \cite{vaswani2017attention}, explicit density models with approximate density like the Variational Autoencoder \cite{kingma2013auto}, and implicit direct density generative models such as Generative Adversarial Networks \cite{goodfellow2014generative}.

\textit{Autoregressive (Fully-observed) generative models} model the observed data directly without introducing dependencies on any new unobserved local variables. Assuming all items in a sequence $x=(x_1, x_2, \ldots, x_N)$ are fully observed, the probability distribution $p(x)$ of the data is modeled in an auto-regressive fashion using the chain rule of probability:

\begin{equation}
p(x_1, x_2, \ldots, x_N)= \prod_{i=1}^{N} p(x_i|x_1, x_2, \ldots, x_{i-1})  
\end{equation}
Training autoregressive models is done by maximizing the data likelihood, allowing these models to be evaluated quickly and exactly. Sampling from autoregressive models is exact, but it is expensive since samples need to be generated in sequential order. Extracting representions from fully observed models is challenging, but this is currently an active research topic.

\textit{Latent variable generative models} explain hidden causes by introducing an unobserved random variable $z$ for every observed data point. The data likelihood $p(x)$ is computed as follows: 

\begin{equation}
    \log p(x) = \int p_{\theta}(x|z) p(z) dz =\mathbb{E}_{p(z)}[p_{\theta}(x|z)]
\end{equation}
Latent models present the advantage that sampling is exact and cheap, while extracting latent features from these models is straightforward. They are evaluated using the lower bound of the log likelihood. 

\textit{Implicit density models} (among which the most famous models are GANs)  introduce a second discriminative model able to distinguish model generated samples from real samples in addition to the generative model. While sampling from these models is cheap, it is inexact. The evaluation of these models is difficult or even impossible to carry, and extracting latent representations from these models is very challenging. We summarize in Table \ref{table_generative_models} characteristics of the three categories of generative models discussed above.

\begin{table}[!htbp]
\caption{Comparison of generative model frameworks.} 
\centering
\begin{tabular}{l | l | l }
\hline
\hline
\textbf{Model type} & \textbf{Evaluation} & \textbf{Sampling}  \\
\hline
\hline
\textbf{Fully-observed} & Exact and & Exact and \\
 & Cheap & Expensive \\
\hline
\textbf{Latent models} & Lower Bound & Exact and  \\
& & Cheap \\
\hline
\textbf{Implicit models} & Hard or & Inexact and  \\
& Impossible & Cheap\\
\hline
\hline
\end{tabular}
\label{table_generative_models}
\end{table}

In what follows we review models for neural language generation from most general to the most specific according to the problem definition categorization presented in Section \ref{problem_definition}; for each model architecture we first list models for generic text generation, then introduce models for conditional text generation, and finally outline models used for constrained text generation. We begin with recurrent neural network models for text generation in Section \ref{recurrent_nn}, then present sequence-to-sequence models in Section \ref{seq2seq_nn}, generative adversarial networks (GANs) in Section \ref{gan_nn}, variational autoencodes (VAEs) in Section \ref{vae_nn} and pre-trained models for text generation in Section \ref{pretrained_nn}. We also provide a comprehensive overview of text generation tasks associated with each model.  

\subsection{Recurrent Architectures}
\label{recurrent_nn}

\subsubsection{Recurrent Models for Generic / Free-Text Generation}

Recurrent Neural Networks (RNNs) \cite{rumelhart1986learning}, \cite{mikolov2010recurrent} are able to model long-term dependencies in sequential data and have shown promising results in a variety of natural language processing tasks, from language modeling \cite{mikolov2012statistical} to speech recognition \cite{graves2013speech} and machine translation \cite{kalchbrenner2013recurrent}. An important property of RNNs is the ability of learning to map an input sequence of variable length into a fixed dimensional vector representation. 

At each timestep, the RNN
receives an input, updates its hidden state, and makes a
prediction. 
Given an input sequence $x=(x_1, x_2, \ldots, x_T)$, a standard RNN computes the hidden vector sequence $h=(h_1, h_2, \ldots, h_T)$ and the output vector sequence $y=(y_1, y_2, \ldots, y_T)$, where each datapoint $x_t, h_t, y_t, \forall \text{ } t \in \{1, \ldots, T\}$ is a real valued vector, in the following way:

\begin{equation}
\begin{split}
    h_t &= \mathcal{H}(W_{xh}x_t + W_{hh}h_{t-1} + b_h) \\
    y_t &= W_{hy}h_t + b_y
\end{split}
\label{rnn_eq}
\end{equation}
In Equation \ref{rnn_eq} terms $W$ denote weight matrices, in particular $W_{xh}$ is the input-hidden weight matrix and $W_{hh}$ is the hidden-hidden weight matrix. The $b$ terms denote bias vectors, where $b_h$ is the hidden bias vector and $b_y$ is the output bias vector. $\mathcal{H}$ is the function that computes the hidden layer representation. Gradients in an RNN are computed via backpropagation through time \cite{rumelhart1986learning}, \cite{werbos1989backpropagation}.
By definition, RNNs are inherently deep in time considering that the hidden state at each timestep is computed as a function of all previous timesteps. While in theory RNNs can make use of information in arbitrarily long sequences, in practice they fail to consider context beyond the few previous timesteps due to the vanishing and exploding gradients \cite{bengio1994learning} which cause gradient descent to not be able to learn long-range temporal structure in a standard RNN. Moreover, RNN-based models contain millions of parameters and have traditionally been very difficult to train, limiting their widespread use \cite{sutskever2011generating}. Improvements in network architectures, optimization techniques and parallel computation have resulted in  recurrent models learning better at large-scale \cite{lipton2015critical}. 

Long Short Term Memory (LSTM)  \cite{hochreiter1997long} networks are introduced to overcome the limitations posed by vanishing gradients in RNNs and allow gradient descent to learn long-term temporal structure. The LSTM architecture largely resembles the standard RNN architecture with one hidden layer, and each hidden layer node is modified to include a memory cell with a self-connected recurrent edge of fixed weight which stores information over long time periods. 
A memory cell $c_t$ consists of a node with an internal hidden state $h_t$ and a series of gates, namely an input gate $i_t$ which controls how much each LSTM unit is updated, a forget gate $f_t$ which controls the extent to which the previous memory cell is forgotten, and an output gate $o_t$ which controls the exposure of the internal memory state.  The LSTM transition equations at timestep $t$ are:

\begin{equation}
\begin{split}
    i_t &= \sigma (W^{(i)}x_t + U^{(i)}h_{t-1} + b^{(i)}) \\
    f_t &= \sigma (W^{(f)}x_t + U^{(f)}h_{t-1} + b^{(f)}) \\
    o_t &= \sigma (W^{(o)}x_t + U^{(o)}h_{t-1} + b^{(o)}) \\
    u_t &= \sigma (W^{(u)}x_t + U^{(t)}h_{t-1} + b^{(t)}) \\
    c_t &= i_t \odot u_t + f_t \odot c_{t-1} \\
    h_t &= o_t \odot \text{tanh}(c_t)
\end{split}
\label{lstm_eq}
\end{equation}
In Equation \ref{lstm_eq}, $x_t$ is the input at the current timestep $t$, $\sigma$ denotes the logistic sigmoid function and $\odot$ denotes elementwise multiplication. $U$ and $W$ are learned weight matrices. LSTMs can represent information over multiple time steps by adjusting the values of the gating variables for each vector element, therefore allowing the gradient to pass without vanishing or exploding. In both RNNs and LSTMs the data is modeled via a fully-observed directed graphical model, where the distribution over a discrete output sequence $y=(y_1, y_2, \dots, y_T)$ is decomposed into an ordered product of conditional distributions over tokens:
\begin{equation}
P(y_1, y_2, \dots, y_T) = P(y_1)\prod_{t=1}^{T}P(y_t|y_1, \dots, y_{t-1})
\end{equation}

Similar to LSTMs, Gated Recurrent Units (GRUs) \cite{cho2014learning} learn semantically and
syntactically meaningful representations of
natural language and have gating units to modulate the flow of information. Unlike LSTMs, GRU units do not have a separate memory cell and present a simpler design with fewer gates. The activation $h_{t}^{j}$ at timestep $t$ linearly interpolates between the activation at the previous timestep $h_{t}^{j-1}$ and the candidate activation $ \widetilde{h}_{t}^{j}$. 
The update gate $z_t^j$ decides how much the current unit updates its content, while the reset gate $r_t^j$ allows it to forget the previously computed state. The GRU update equations at each timestep $t$ are:

\begin{equation}
\begin{split}
h_{t}^{j} &= (1-z_t^j)h_{t-1}^{j} + z_t^j \widetilde{h}_{t}^{j}\\
z_t^j &= \sigma(W_z x_t + U_z h_{t-1})^j \\
\widetilde{h}_{t}^{j} &= \tanh (Wx_t + U(r_t \odot h_{t-1}))^j \\
r_t^j &= \sigma(W_r x_t + U_r h_{t-1})^j
\end{split}
\label{eq_gru}
\end{equation}

Models with recurrent connections are trained with \textit{teacher forcing} \cite{williams1989learning}, a strategy emerging from the maximum likelihood criterion designed to keep the recurrent model predictions close to the ground-truth sequence.
At each training step the model generated token $\hat{y}_t$ is replaced with its ground-truth equivalent token $y_t$,  
while at inference time each token is generated by the model itself (i.e. sampled from its conditional distribution over the sequence given the previously generated samples).
The discrepancy between training and inference stages leads to exposure bias, causing errors in the model predictions that accumulate and amplify quickly over the generated sequence,
\cite{lamb2016professor}. As a remedy, \textit{Scheduled Sampling} \cite{bengio2015scheduled} mixes inputs from the ground-truth sequence with inputs generated by the model at training time, gradually adjusting the training process from fully guided (i.e. using the true previous token) to less guided (i.e. using mostly the generated token) based on curriculum learning \cite{bengio2009curriculum}.
While the model generated distribution can still diverge from the ground truth distribution as the model generates several consecutive tokens, possible solutions are:
\textit{i)} make the self-generated sequences short, and \textit{ii)} anneal the probability of using self-generated vs. ground-truth samples to 0, according to some schedule. Still, models trained with scheduled sampling are shown to memorize the distribution of symbols conditioned on their position in the sequence instead of the actual prefix of preceding symbols \cite{huszar2015not}. 



Many extensions of vanilla RNN and LSTM architectures are proposed in the literature aiming to improve  generalization and sample quality \cite{yu2019review}. Bidirectional RNNs \cite{schuster1997bidirectional}, \cite{berglund2015bidirectional} augment unidirectional  recurrent models by introducing a second hidden layer with connections flowing in opposite temporal order to exploit both past and future information in a sequence. Multiplicative RNNs \cite{sutskever2011generating} allow flexible input-dependent transitions, 
however many complex transition functions hard to bypass. 
Gated feedback RNNs and LSTMs \cite{chung2014empirical} rely on gated-feedback connections to enable the flow of control signals from the upper to lower recurrent layers in the network. Similarly, depth gated LSTMs \cite{yao2015depth} introduce dependencies between lower and upper recurrent units by using a depth gate which connects memory cells of adjacent layers. Stacked LSTMs stack multiple layers at each time-step to increase the capacity of the network, while nested LSTMs \cite{moniz2018nested} selectively access LSTM memory cells with inner memory. Convolutional LSTMs  \cite{sainath2015convolutional}, \cite{xingjian2015convolutional} are designed for jointly modeling spatio-temporal sequences. Tree-structured LSTMs \cite{zhu2015long}, \cite{tai2015improved} extend the LSTM structure beyond a linear chain to tree-structured network topologies, and are useful at semantic similarity and sentiment classification tasks. Multiplicative LSTMs \cite{krause2016multiplicative} combine vanilla LSTM networks of fixed weights with multiplicative RNNs to allow for flexible input-dependent weight matrices in the network architecture. 
Multiplicative Integration \cite{wu2016multiplicative} RNNs achieve better performance than vanilla RNNs by using the Hadamard product 
in the computational additive building block of RNNs. Mogrifier LSTMs \cite{melis2019mogrifier} capture interactions between inputs and their context by mutually gating the current input and the previous output of the network.
For a comprehensive review of RNN and LSTM-based network architectures we point the reader to \cite{yu2019review}. 


\subsubsection{Recurrent Models for Conditional Text Generation}

A recurrent free-text generation model becomes a conditional recurrent text generation model when the distribution over training sentences is conditioned on another modality. For example in machine translation the distribution is conditioned on another language, in image caption generation the condition is the input image, in video description generation we condition on the input video, while in speech recognition we condition on the input speech. 

Content and stylistic properties (such as sentiment, topic, style and length) of generated movie reviews are controlled in a conditional LSTM language model by conditioning on context vectors that reflect the presence of these properties \cite{ficler2017controlling}. Affective dialogue responses are generated by conditioning
on affect categories in an LSTM language model \cite{ghosh2017affect}. A RNN-based language model equipped with dynamic memory outperforms more complex memory-based models for dialogue generation \cite{mei2017coherent}. Participant roles and conversational topics are represented as context vectors and incorporated into a LSTM-based response generation model \cite{luan2016lstm}.  

\subsubsection{Recurrent Models for Constrained Text Generation}

Metropolis-Hastings
sampling \cite{miao2019cgmh} is proposed for both soft and hard constrained sentence generation from models based on recurrent neural networks. The method is based on Markov Chain Monte Carlo (MCMC) sampling and performs local operations such as insertion, deletion and replacement in the sentence space for any randomly selected word in the sentence. 

Hard constraints on the generation of scientific paper titles are imposed by the use of a forward- backward recurrent language model which generates both previous and future words in a sentence conditioned on a given topic word \cite{mou2015backward}. While the topic word can occur at any arbitrary position in the sentence, the approach can only generate sentences constrained precisely on one keyword. Multiple constraints are incorporated in sentences generated by a backward-forward LSTM language model by lexically substituting  constrained tokens with their closest matching neighbour in the embedding space \cite{latif2020backward}.
Guiding the conversation towards a designated topic while integrating specific vocabulary words is achieved by combining discourse-level rules with neural next keywords prediction \cite{tang2019target}. A recurrent network based sequence
classifier is used for extractive summarization in \cite{nallapati2017summarunner}. Poetry generation which obeys hard rhythmic,
rhyme and topic constraints is proposed in \cite{ghazvininejad2016generating}.

\subsection{Sequence-to-Sequence Architectures}
\label{seq2seq_nn}

Although the recurrent models presented in Section \ref{recurrent_nn} present good performance whenever large labeled training sets are available, they can only  be applied to problems whose inputs and targets are encoded with vectors of fixed dimensionality. Sequences represent a challenge for recurrent models since RNNs require the dimensionality of their inputs and
outputs to be known and fixed beforehand. In practice, there are many problems in which the sequence length is not known a-priori and it is  necessary to 
map variable length sequences into
fixed-dimensional vector representations. To this end, models that can map sequences to sequences are proposed. These models makes minimal assumptions on the sequence structure and learn to map an input sequence into a vector of fixed dimensionality and then map that vector back into an output sequence, therefore learning to decode the target sequence from the encoded vector representation of the source sequence. We present these models in detail below.


\subsection{Sequence-to-sequence models that condition on the input text}


Sequence-to-sequence (seq2seq) models \cite{kalchbrenner2013recurrent}, \cite{sutskever2014sequence}, \cite{cho2014learning} are conditional language models which can deal with variable length inputs and output. Also known as encoder-decoder models, they have been very successful in machine translation \cite{luong2015effective}, text summarization \cite{nallapati2016abstractive}, dialogue systems \cite{vinyals2015neural}, and image and video captioning \cite{vinyals2015show}. Seq2seq models consist of two paired recurrent neural networks: the first network (encoder) summarizes a variable-length source sequence of symbols $x=(x_1, x_2, \ldots, x_T)$ into a rich fixed-length vector representation $v$, while the second network (decoder) uses the vector representation $v$ as its initial hidden state and deciphers it into another variable-length target
sequence of symbols $y=(y_1, y_2, \ldots, y_{T'})$ by computing the conditional probability of each word $y_t$ in the target sequence
given the previous words $y_1, y_2, \ldots, y_{t-1}$ and the input $x$, where length $T$ of input may differ from length $T'$ of output. The conditional probability $p(y_1, y_2, \ldots, y_{T'}|x_1, x_2, \ldots, x_T)$ of the output sequence $y$ given the input sequence $x$ can be formally expressed as:

\begin{equation}
\begin{split}
    p(y_1, y_2, \ldots, y_{T'}|x_1, x_2, \ldots, x_T) = \\ \prod_{t=1}^{T'}p(y_t|v, y_1, y_2, \ldots, y_{t-1})
\end{split}
\label{eq_seq2seq}
\end{equation}
Typical architectural choices for the encoder and decoder are RNN and LSTM neural networks. In addition, deep convolutional networks are proposed to model long-range dependencies in lengthy documents \cite{dauphin2017language}, \cite{gehring2017convolutional}. Given training set consisting of $N$ paired input and output sentences, a sequence-to-sequence model is trained using maximum likelihood to maximize the conditional log-likelihood of the correct target sentence $y_n$ given the source sentence $x_n$ for every ($x_n$, $y_n$) pair and model parameters $\theta$:

\begin{equation}
    \max_{\theta} \frac{1}{N}\sum_{t=1}^{N}\log p_{\theta}(y_n|x_n)
\label{eq_seq2seq_training}
\end{equation}
After training, the sequence-to-sequence model can be used in two ways: \textit{i)} to produce a probability score $p_{\theta}(y|x)$ for a given pair $(x,y)$ of input $x$ and output $y$ sequences, or  \textit{ii)} to generate the target sequence $y$ corresponding to the input sequence $x$. In the latter case, decoding methods are used to generate the most likely output sequence. In what follows we present decoding strategies for neural sequence-to-sequence generative models.

\paragraph{Decoding}  Generating the most likely output sequence from a trained model involves running an exhaustive search over all possible output sequences, scoring them based on their likelihood and selecting the most likely sequence $\hat{y}$ such that:

\begin{equation}
    \hat{y}= \arg \max_y P(y|x) = \arg \max_y \prod_{t=1}^{N} P(y_t|y_{<t}, x)
\label{eq_max_likelihood_seq}
\end{equation}
In Equation \ref{eq_max_likelihood_seq}, the model outputs a probability distribution over the next token at each timestep $t$ given the input $x$ and the previously predicted tokens $y_{<t}$. The probability distribution over the next word in the target vocabulary $w_i \in V$ is commonly modeled using a softmax function:

\begin{equation}
    P(y_t = w_i|y_{<t}, x) = \frac{\exp(z_{t, i})}{\sum_{j=1}^{V}\exp(z_{t, j})}, 
\end{equation}
where $z_t=f(y_{<t},x)$ represents the output of the encoder-decoder model given input sequence $x$ and the sequence of tokens predicted so far $y_{<t}$. While Equation \ref{eq_max_likelihood_seq} theoretically outputs the optimal output sequence $\hat{y}$, in practice it is intractable to run an exhaustive search to find $\hat{y}$ precisely. The exact decoding problem is exponential in the length of the source sentence $x$, and factors such as the branching factor, number of timesteps and the large vocabulary size impede yielding precisely the most probable sequence $\hat{y}$ from the trained model. Alternative decoding strategies based on heuristic search are used to find reasonable approximations of the optimal output sequence. As opposed to exact decoding, these sampling techniques are incomplete decoding strategies which exclude tokens from consideration at each step, and generate a random sequence according to the learnt  probability distribution. The decoding strategy of choice bears a huge impact on the quality of the generated machine text, even when the same neural language model is used for generation \cite{holtzman2019curious}. In what follows we present decoding methods commonly used in the NLG literature, noting that the best decoding strategy for text generation from a trained language model is still largely an unresolved problem.

\subparagraph{Argmax/  Greedy search} The argmax sampler is  the simplest approach to decoding a
likely sequence. At each timestep it greedily selects the most likely (argmax) token over the softmax output distribution and feeds it as input to the next timestep until the end-of-sentence token is reached, as follows:

\begin{equation}
    \hat{y}_t = \arg \max_{y_t}P(y_t|y_{<t}, x)
\end{equation}
Greedy decoding preserves a single hypothesis at each timestep, however selecting the best individual token output per timestep does not necessarily result in the best overall output hypothesis -- there may well exist a better path which includes a less likely token (not argmax) at some point in the decoding process; the method will also miss a high-probability token hiding after a low-probability one. In addition, other limitations of greedy decoding include the generation of repetitive and short output sequences which lack in diversity even for large well-trained models \cite{holtzman2019curious}, \cite{radford2019language}, and the impossibility to generate multiple samples during decoding. This makes it a suboptimal decoding strategy \cite{chen2018stable}. 

\subparagraph{Random/ Stochastic/ Temperature search/ Ancestral Sampling} Stochastic sampling introduces randomness in  decoding and arbitrarily samples each output token from the model's distribution at each timestep. A temperature parameter $T>0$ is commonly used to control how flat ($T \rightarrow \infty$) or greedy ($T \rightarrow 0$) the multinomial distribution over the next token is \cite{ackley1985learning}. Values of $T>1$ cause increasingly more random outputs, while $T \approx 0$ resembles greedy sampling:

\begin{equation}
    P(y_t=w_i|y_{<t}, x) = \frac{\exp(z_{t,i}/T)}{\sum_{j=1}^V \exp(z_{t,j}/T)}
\label{random_sampling}
\end{equation}

It is important to note that doing completely random sampling can negatively impact sequence generation as it introduces unlikely words and mistakes not encountered at training time \cite{fan2018hierarchical}. 

\subparagraph{Beam search} Beam search is an approximate graph-based inference algorithm which explores the hypothesis space in a greedy left-to-right (breadth-first) manner over a limited portion of the overall search space. Each hypothesis is expanded iteratively one token at a time and only the $k$-best hypotheses are eventually kept, where $k$ denotes the beam width or beam size. Unlike greedy decoding which maintains a single hypothesis at a time and can miss out a highly probable token when it is preceded by a low probable one, beam search explores multiple sequences in parallel and mitigates the aforementioned problems by maintaining a beam (set of $k$ hypotheses) of potential sequences constructed word-by-word. More specifically, the decoding process begins with the start-of-sentence token and at every step of decoding new $k$-best tokens $w_i$ are selected according to the probability distribution $P(y_t=w_i|y_{<t},x)$. Each partial hypothesis is expanded with a new token and its cumulative log-probability is updated accordingly to capture model's preferences. The process repeats until the end-of-sentence token is produced, at which time the hypothesis is complete. Finally, all complete hypotheses are scored in descending order of their likelihood and only the $k$-best scoring hypothesis are preserved. Decoding new sequences from the trained model is equivalent to finding the sequence $y^*$ 
that is most probable under the model distribution:

\begin{equation}
  y^* = \arg \max_{y} p(y|x) = \arg \min_{y} - \log p(y|x)
\end{equation}

Beam search was effective in early work on neural machine translation \cite{sutskever2014sequence} and has become the standard algorithm for many language generation tasks at sampling sufficiently likely sequences from  probabilistic encoder-decoder models. Nevertheless, beam search is sensitive to output length and best results are obtained when the length of the target sentence is predicted before decoding \cite{murray2018correcting}, \cite{yang2018breaking}. Beam search decoding is also slow as it introduces a substantial computational overhead \cite{cho2016noisy} and the candidate sequences it produces are short, dull, generic, and include common phrases and repetitive text from the training set \cite{shao2016generating}, \cite{vijayakumar2016diverse}, \cite{sordoni2015neural}, \cite{li2016diversity}, \cite{wolf2019transfertransfo}. While the use of maximum likelihood as a training objective leads to high quality models for many language understanding tasks, maximization based decoding results in neural text degeneration \cite{holtzman2019curious}. In addition, the likelihood objective assigns too much probability to repetitive and frequent words, focusing only on producing the next word and not on optimizing sequence generation \cite{welleck2019neural}. Consequently, the generated outputs produced by beam search lack in diversity \cite{li2016simple}, \cite{li2016diversity}, \cite{gimpel2013systematic} and are largely variations of the same high likelihood beam with minor differences in punctuation and morphology \cite{li2016mutual}.   Increasing the beam size leads to a degrade in performance and negatively affects sequence generation quality \cite{koehn2017six}, \cite{yang2018breaking}. As an alternative to maximum likelihood training, unlikelihood training is proposed to force the model to assign lower probability scores to unlikely
generations \cite{welleck2019non}.

Extensions of beam search are proposed focusing on improving output diversity, and can be applied either during the decoding process or post-decoding. We present diversity promoting methods used during the decoding process first. \textit{$N$-gram blocking} \cite{paulus2017deep}, \cite{klein2017opennmt} discards a hypothesis if the occurrence frequency of any token within it is greater than one; this strategy is used to block previously generated $n$-grams from subsequent generation. \textit{Iterative beam search} \cite{kulikov2019importance} runs multiple iterations of beam search on disjunct areas of the search space, ensuring there is no overlap between the current search space and areas explored by previous iterations. \textit{Diverse beam search} \cite{vijayakumar2016diverse} augments beam search with a diversity promoting term which ensures a candidate hypothesis is sufficiently different from other partial
hypotheses according to standard diversity functions such as $n$-gram diversity, neural embedding diversity and Hamming distance. \textit{Cluster-based beam search} \cite{tam2020cluster} clusters semantically similar partial hypothesis using $k$-means clustering, followed by hypothesis pruning to keep only top candidate hypotheses from each cluster. \textit{Noisy parallel decoding} \cite{cho2016noisy} can be combined with any decoding strategy and works by injecting noise (randomly sampled from a normal distribution) into the hidden state of the decoder at each timestep, followed by running in parallel multiple approximate decoding processes. \textit{Top-g capping beam search} \cite{li2016simple}, \cite{li2016mutual} incentives diversity by grouping candidate hypotheses according to their parent, and selects top-$g$ candidates from each group. Post-decoding diversity-promoting methods are also proposed. The simplest strategy to increase the diversity of outputs is to cluster decoded sentences and remove highly similar candidates \cite{kriz2019complexity}. Along the same line, it is possible to over-sample generated sequences  from the model and filter them down to retain a smaller number of outputs
\cite{ippolito2018comparison}; random sampling is the recommended way to over-sample candidates.


Lexically constrained decoding with \textit{grid beam search} is used to enforce lexical constraints (words or phrases) and incorporate additional knowledge in the generated output \cite{hokamp2017lexically}. The search space is hard constrained to produce only candidates which
contain one or more pre-specified sub-sequences. Similarly,  \textit{constrained beam search} \cite{anderson2017guided} is used to force the inclusion of specific  words in the generated sequences by adopting a finite-state machine approach which recognizes valid  constrained and complete outputs.  \textit{Dynamic beam allocation} \cite{post2018fast} improves the time efficiency of constrained decoding by grouping beam candidates according to how many constraints they meet. \textit{Vectorized dynamic beam allocation} achieves even faster decoding by organizing into a trie the constraints that
have not yet been generated  \cite{hu2019improved}. 

In addition, decoding methods that optimize for output with high probability produce incoherent, repetitive and generic output sequences. Beam search and greedy decoded texts fail to reproduce the distribution of words in human generated texts \cite{holtzman2019curious}.  As a remedy, decoding strategies that truncate the neural probability distribution at different thresholds establish a trustworthy prediction zone from which tokens can be sampled according to their relative probabilities. Next we present thresholding-based decoding strategies. 



\subparagraph{ Top-k sampling} The top-$k$ random sampling  \cite{fan2018hierarchical} scheme restricts sampling from the $k$ most likely terms in the distribution and introduces randomness in the decoding process. New words are generated at each timestep  by randomly selecting $k$ (typically $k=10$) tokens from the most likely candidate tokens sampled from the probability distribution of each word in the vocabulary being the likely next word given the previously selected words. This decoding scheme is found more effective than beam search \cite{radford2019language}, \cite{holtzman2018learning}.

\subparagraph{ Nucleus (Top-p) sampling} Nucleus sampling \cite{holtzman2019curious} suppresses the unreliable tail of the probability distribution consisting of tokens with relatively low probability and samples tokens from the remaining top-$p$ portion or nucleus, which concentrates highest probability mass. This approach  allows the model to generate tokens from the vast majority of the probability mass and prevents sampling low probability tokens.

\subparagraph{Penalized sampling} Penalized sampling \cite{keskar2019ctrl} is used at inference time to encourage output diversity. By discounting the probability of the already generated tokens in a sequence, the model is discouraged from generating them again.

\paragraph{Sequence Generation} In what follows we present the dominant approaches used in neural sequence generation.

\begin{enumerate}
	\item \textbf{Monotonic / Autoregressive Sequence Generation}: In natural language generation sequences are typically generated iteratively following a left-to-right generation order in which new tokens are added successively  to the end of an unfinished sequence. Monotonic neural text generation decomposes the sequence prediction problem into a series of next token predictions, i.e. given input sequence $x$ and assuming the availability of ground-truth  previous tokens $(y_1^*,y_2^*, \ldots, y_{t-1}^*)$  predict the next token $y_t$: 
	
	\begin{equation}
	    P(Y|x) = \prod_{t=1}^T p(y_t| y_1, y_2, \ldots, y_{t-1}, x) 
	\end{equation}
	
	The learning process maximizes the log-probability  of a correct output sequence given the input $x$, teaching the model to predict the correct next token $p(y_t^*| y_1^*, y_2^*, \ldots, y_{t-1}^*, x)$ by using a cross-entropy loss applied at each decoding step:
	
	\begin{equation}
	    \max \log p(Y^*|x) = \sum_{t=1}^{T} \log p(y_t^*| y_1^*, y_2^*, \ldots, y_{t-1}^*, x)
	\end{equation}
	
	With recent advances in text representation learning, monotonic generation of sequences has become the standard approach in neural text generation. Autoregressive  models are easy to train and achieve robust performance on large datasets. To speed up their training, models that leverage parallelism at training time replace recurrent layers in the decoder with masked convolution layers \cite{kalchbrenner2016neural}, \cite{gehring2017convolutional} or self-attention \cite{vaswani2017attention}.  Nevertheless, at inference time autoregressive decoding with beam search can be slow as the individual steps of the decoder must run sequentially  \cite{gu2017non}.
	
	\item \textbf{Parallel Sequence Generation}: Recent work in text generation is challenging the assumption that text needs to be generated sequentially \cite{gu2017non}. Indeed, the simplistic procedure of sequential token generation does not reflect how humans write text \cite{guu2018generating} and is limiting content diversity \cite{mehri2018middle}. In contrast to standard autoregressive models that predict each word conditioned on all previous words and naturally model the length of a target sequence, non-autoregressive models enable parallel generation of output tokens and incorporate target sequence length prediction at inference time \cite{lee2018deterministic}. Sequence generation in parallel speeds up inference by leveraging parallel computation, and captures dependencies between tokens by iteratively refining a sequence \cite{lee2018deterministic}.
	Parallel decoding models include iterative refinement \cite{lee2018deterministic}, noisy parallel decoding \cite{gu2017non}, masked language models \cite{ghazvininejad2019mask}, \cite{ghazvininejad2020semi}, insertion-based methods \cite{stern2019insertion}, \cite{chan2019kermit}, \cite{gu2019insertion}, edit-based methods \cite{gu2019levenshtein}, \cite{ruis2020insertion}, normalizing flow models \cite{ma2019flowseq} and connectionist temporal classification \cite{libovicky2018end}. 
	
	Non-autoregressive generation models approach the performance of autoregressive models and have been successfully applied in machine translation \cite{gu2017non},
	\cite{guo2019non}, \cite{saharia2020non} and speech synthesis \cite{oord2018parallel}. Nevertheless, they make the limiting assumption that output tokens are conditionally independent given the input, which leads to the presence of redundant tokens in the non-autoregressive generated sequences. In addition, unlike their autoregressive counterparts which stop generation by emitting the end-of-sentence token, non-autoregressive models need to explicitly incorporate output length prediction as a preliminary generation step.    
	
	
	\item \textbf{Non-Monotonic Sequence Generation}: As opposed to monotonic sequence generation, 
	flexible sequence generation produces an output sentence without following a strict pre-defined left-to-right order. To this end, non-monotonic generation approaches decompose a ground-truth sequence $Y$ into a multi-set of items to be generated $\mathcal{Y}$ and a set of ordering constraints $\mathcal{C}$ \cite{welleck2018loss}. Naturally, the order is which sequences are generated impacts performance \cite{vinyals2015order}. Generating sequences in arbitrary orders by simultaneously predicting a word and the position in which it should be inserted during each decoding step presents comparable performance to conventional left-to-right generation \cite{gu2019insertion}. A hierarchical approach to decoding is proposed by deliberation networks  \cite{xia2017deliberation}, where the first-pass decoder generates a raw sequence which is then further polished and refined by a second decoder. Review networks \cite{yang2016review} further edit the encoder hidden states before generating the output sentence. 
\end{enumerate}


Different layers in sequence-to-sequence models exhibit different functionality and learn different representations. While lower layers of the encoder learn to represent word structure, higher layers of the encoder capture semantics and word meaning   \cite{belinkov2017neural}. This is consistent with findings on representations learnt by CNNs on image data \cite{zeiler2014visualizing}.  

Sequence-to-sequence models are trained in a multitask learning settings in which either the encoder, the decoder or both encoder and decoder are shared between multiple tasks \cite{luong2015multi}, \cite{dong2015multi}. 

\paragraph{Attention} The attention mechanism \cite{bahdanau2014neural}, \cite{luong2015effective} is proposed to enhance seq2seq models with a random access memory which allows to handle long input sequences and focus on salient pieces of input information. Attention dynamically attends to different parts of the input while generating
each target-side word. In order to estimate the relevance of input tokens, the distribution of attention weights is computed over all input tokens and higher values are assigned to those tokens considered relevant. The attention mechanism is a crucial component of many seq2seq models used in image captioning \cite{xu2015show}, machine translation \cite{jean2015using}, \cite{luong2015stanford}, constituency parsing \cite{vinyals2015grammar}, visual object tracking \cite{mnih2014recurrent}, abstractive summarization \cite{rush2015neural}, \cite{nallapati2016abstractive}. Besides performance gains, attention is also commonly  used as a tool for
interpreting the behaviour of neural architectures since it allows to dynamically highlight
relevant features of the raw input data \cite{hermann2015teaching} or higher-level neural representations \cite{galassi2019attention}.


In the context of encoder-decoder models for neural machine translation \cite{bahdanau2014neural}, attention is designed  to learn alignments between the decoding states and the encoded memories. Therefore, attention makes all encoder hidden states available to the decoder at decoding time (i.e. soft attention) as opposed to regular seq2seq models where the decoder can only access the last encoder hidden state. The attention mechanism computes alignment weights for all input positions, and decides how much information to retrieve from the input by learning where to focus. The benefit of using attention is that the encoder no longer needs to encode all source-side information into a fixed-length vector, while the decoder can selectively retrieve information spread throughout the entire input sequence. Empirical evidence shows that the attention model is more efficient than the encoder-decoder approach since its dynamic alignment mechanism requires less parameters and training instances \cite{jean2015using}. 

In its basic formulation, the attention function maps a sequence of $K$ vectors or keys $k_i$ with dimensionality $d_k$ corresponding to input features (either word or character level embeddings) to a distribution $a$ of weights $a_i, |a| = d_k$ for the input query $q$. If $q$ is defined (for eg., machine translation, question answering), input elements which are relevant to $q$ will be selected; if $q$ is undefined (for eg., document classification), inherently relevant input elements are selected. The compatibility function $f$ is used to measure how well the query matches the keys, yielding a vector $e$ of energy scores \cite{zhao2018attention} with dimensionality $d_k$ where each element $e_i$ represents the relevance of key $k_i$ to query $q$ under $f$; please see Table \ref{table_compatibility}.

\begin{equation}
\begin{split}
    e &= f(q,K) : \text{energy scores} \\
\end{split}
\end{equation}


\begin{table*}[!h]
\caption{Attention compatibility functions. }
\centering
\small
\begin{tabular}{l| l | l}
\hline
\hline
\textbf{Name} & \textbf{ Alignment function} & \textbf{Reference}\\
\hline
\hline
Similarity / Content-based  & $f(q, K) = \cos (q,k)$ & \cite{graves2014neural} \\
\hline
Additive / Concat & $f(q, K) = v_a^T\tanh(W[K; q]) $ & \cite{bahdanau2014neural},\\
& & \cite{luong2015effective} \\
\hline
General / Bilinear & $f(q, K) = q^T W K$  & \cite{luong2015effective} \\
\hline
Dot-Product & $f(q, K) = q^T K$ & \cite{luong2015effective} \\
\hline
Scaled Dot-Product & $f(q, K) = \frac{q^T K}{\sqrt(d_k)}$ & \cite{vaswani2017attention} \\
\hline
Location-based & $f(q, K) = \text{softmax}(Wq)$ & \cite{luong2015effective} \\
\hline
\hline
\end{tabular}
\label{table_compatibility}
\end{table*}

Next, the energy scores $e$ are transformed into a vector $a$ of attention weights $a_i$ with dimensionality $d_k$ by mapping to the distribution function $g$ (softmax function is a common choice). While the attention weights $a_i$ still represent the relevance of each element $k_i$ to the query, new representations of the keys $k_i$ are computed under the form of sequence $V$ of $d_k$ vectors $v_i$ \cite{cui2017attention}. There is a one-to-one mapping between elements of $V$ and $K$, and the two vectors are different representations of the same data. Nevertheless, attention weights $a_i$ are applied on vectors values $v_i$ to obtain vector $Z$ of attention-weighted representations of $V$. Finally, all elements $z_i$ of $Z$ are aggregated to obtain a compact representation of the input in the form of context vector $c$:

\begin{equation}
\begin{split}
    a &= g(e) : \text{attention weights} \\
    z_i &= a_i v_i : \text{weighted representations} \\
    c &= \sum_{i=1}^{d_k} z_i : \text{context vector} \\
\end{split}
\end{equation}

In the literature attention mechanisms are categorized based on whether attention is placed on all or just a few source positions. In what follows we review the main categories of attention models. 



 




\paragraph{Soft vs. Hard Attention} The distinction between soft and hard attention is proposed in image caption generation  \cite{xu2015show}, based on whether the attention model has access to the entire image or just an image patch. \textit{Deterministic Soft Attention} places the attention weights ``softly'' over all patches in the source image. The model is differentiable and can be trained via standard back-propagation. Nevertheless, it can be expensive to compute when the source input is large. \textit{Stochastic Hard Attention} only selects a one patch of the image  to attend at a time. While at inference time hard attention is less expensive to compute  compared to soft attention, it is non-differentiable. To this end, hard attention is trained either by maximizing an approximate variational lower bound, or via the Reinforce \cite{williams1992simple} algorithm. 

\paragraph{Global vs. Local Attention} The distinction between global and local attention is proposed in the context of machine translation \cite{luong2015effective}. \textit{Global Attention} attends to all source words for each target word, i.e. all hidden states of the encoder are used to calculate the context vector $c_t$ as the weighted average over all source states according to attention values $a_t$. Global attention is same as the deterministic soft attention proposed in \cite{xu2015show} and resembles the attention mechanism in \cite{bahdanau2014neural} with minor architectural differences. 
Nevertheless, since global attention simultaneously attends to all words on the source side for each target word, it is computationally expensive and impractical in scenarios where the source sentence is long. \textit{Local attention} considers for each target word only a subset of source words from a small context window at a time.  Local attention combines the soft and hard attention mechanisms proposed in \cite{xu2015show} -- it eliminates the extensive computational needs of soft attention and adds differentiability to hard attention. For the current target word at time $t$, the model identifies a single aligned source position $p_t$ by either assuming source-to-target monotonic alignments or by predicting it. The context window of words $[p_t -D, p_t + D], D 
\in \mathbb{N}^*$ centered at $p_t$ is then used to compute the context vector $c_t$. 

\paragraph{Self-Attention  / Intra-Attention} Aiming to discover lexical relations between tokens in an input sequence, memory and attention are combined within a sequence encoder to create an attention-based memory addressing
mechanism which can generate contextual representations of input tokens \cite{cheng2016long}. The intra-attention mechanism can either be used for single sentences to compute a sentence representation which relates different positions in the sequence, or integrated with encoder-decoder architectures to identify unidirected (and presumably latent) relations between input tokens which mimic the human memory span. All intermediate relations captured by self-attention are soft and differentiable.  

Self-attention was first applied in the context of machine reading \cite{cheng2016long}, where a LSTM architecture is enhanced with a memory network \cite{weston2014memory} to extract and represent meaning from natural language text. 
Self-attention is a general mechanism that can be applied to a wide variety of network architectures and tasks; it can be used as a stand-alone layer and is especially effective when used in later layers \cite{ramachandran2019stand}. Self-attention has become increasingly popular in recent years in a variety of tasks including reading comprehension, abstractive summarization, question answering, textual entailment, learning task-independent sentence representations, and is an integral component of many state-of-the-art neural network models \cite{radford2019language}, \cite{devlin2018bert}. 

\paragraph{Multi-head Attention}

Nevertheless, a single attention layer, especially when computed as a simple weighted average, 
cannot model complex functions. As opposed to self-attention which performs a single attention function at a time, multi-head attention consists of several
attention layers (or  ``heads'') running in parallel and focusing simultaneously on different parts of the input. These attention heads jointly attend to information from different representation subspaces at different positions. Models consisting entirely of multi-headed attention have led to considerable progress on a diverse range of language processing tasks, and in many cases have successfully replaced the more complex recurrence or convolutional neural mechanisms.

Transformer \cite{vaswani2017attention} is the first transduction model based exclusively on attention mechanisms which is highly paralellizable and can handle long-term dependencies while entirely omitting recurrent and convolutional layers. 
The model consists of a stacked encoder-decoder architecture with self-attention \cite{cheng2016long} and point-wise, fully
connected layers in both the encoder and decoder. The encoder embodies a stack of six identical layers, where each encoder layer contains two sub-layers: \textit{i)} a multi-head self-attention mechanism, followed by \textit{ii)} a position-wise fully connected feed-forward network. Similarly, the decoder is also composed of a stack of six layers, where each decoder layer contains three sub-layers: \textit{i)} a multi-head
attention over the output of the encoder stack, \textit{ii)} a multi-head self-attention mechanism, and \textit{iii)} a position-wise fully connected feed-forward network. In addition, to prevent incorporating any future information at decoding time and keep the model auto-regressive, causal constraints are placed on the self-attention decoder blocks. Finally, in the absence of any position-aware recurrence or convolutional mechanisms, sequence ordering information is provided to the model via relative or absolute positioning encodings injected into the input embeddings at the
bottom of the encoder and decoder stacks.  

Transformer is the first entirely
attention-based model applied to machine translation. Improvements in the memory and computational efficiency of the model are proposed in numerous follow-up works.  
Notably, the on-going trend nowadays is to extend the Transformer architecture to  Transformer-based models larger than ever before, and train them on datasets bigger than ever before with  superior performance on various sequence learning tasks, including neural machine translation, language understanding, and
sequence prediction. We present these models in more detail Section \ref{pretrained_nn}.

\subsubsection{Sequence-to-sequence models that handle additional conditions}

Sequence to sequence \cite{sutskever2014sequence} models can be conditioned on specific attributes at training time so as to control their output at inference time. In the literature  the main ways in which generative encoder-decoder models are conditioned are categorized \cite{sennrich2016controlling} as follows: \textit{i)} adding special tokens at the beginning \cite{johnson2017google} or end \cite{sennrich2016controlling} of the source text , \textit{ii)} incorporating additional conditions into the decoder hidden states, therefore bypassing attention \cite{logeswaran2018content}, and \textit{iii)} connecting the conditions directly to the decoder output layer. In addition, when categorical attributes are used in conjuction with end-to-end neural  text classification models, incorporating these attributes in the attention mechanism is the least effective method  \cite{amplayo2019rethinking}. 

Attribute-conditioned review generation uses an attention-enhanced attribute-to-sequence model to generate reviews conditioned on specific product attributes \cite{dong2017learning}, \cite{tang2016context}. Natural language descriptions of database events are generated via encoder-decoder models for concept-to-text generation \cite{mei2016talk} and table-to-text generation  \cite{liu2018table}. Generating questions from long documents is achieved by combining a sequence-to-sequence model with multi-stage attention to represent the broad document context \cite{tuan2019capturing}. A hierarchical sequence-to-sequence architecture is proposed for generating dialogue responses by first sampling a continuous variable in the latent space and then generating the response conditioned on that latent variable \cite{serban2017hierarchical}. A dual attention sequence-to-sequence framework which conditions on both the source text and factual information is used in abstractive summarization for encouraging faithfulness of the generated content to the source document   \cite{cao2018faithful}.

\subsubsection{Sequence-to-sequence models that handle constraints}

Hierarchical story generation systems \cite{fan2018hierarchical} use a two-step approach to text generation: first generate a prompt describing the topic of the story, and then  conditioned on the given prompt generate the content of the story. To this end, fusion mechanisms \cite{sriram2018cold} used on top of sequence-to-sequence models encourage conditioning on the story outline and are found useful at building dependencies between the given inputs and the generated outputs. Entity-focused story generation is proposed in \cite{clark2018neural}. 

Concise summaries of a specific desired length are obtained by controlling the output sequence length for neural encoder-decoder models through either learning or decoding-based methods \cite{kikuchi2016controlling}. The most salient sentences in a document are identified using extractive summarization and then paraphrased using an encoder-decoder based sentence abstractor model \cite{nikolov2020abstractive}. Nevertheless, in abstractive summarization conventional sequence-to-sequence models often suffer from repetition and semantic irrelevance \cite{lin2018global}, and the attention-based encoder outputs are noisy when there is no obvious alignment relationship between the source
text and the target summary  \cite{zhou2017selective}. To alleviate the problem, global encoding of the source context is used to filter encoder outputs and refine represnetations learnt at each timestep based on the global context \cite{lin2018global}. An encoder-decoder framework for extractive  summarization  which incorporates both a sentence encoder and a document encoder to capture context surrounding a sentence, as well as a document decoder to predict sentence labels for inclusion in the summary
based on representations learned by the document encoder is proposed in \cite{zhang2018neural}. A hierarchical document
encoder is combined with an attention-based extractor for selecting sentences and words in extractive summarization \cite{cheng2016neural}. Other approaches to hierarchical  abstractive multi-document summarization combine maximal marginal relevance which serves to extract sentences from the original document based on their relevance and redundancy with the pointer-generator network \cite{see2017get} used to alternate between copying words from the source documents with outputting other vocabulary words \cite{fabbri2019multi}. In text simplification appending special tokens to input sentences at training time is found to improve performance \cite{nishihara2019controllable}. Similarly, sequence to sequence models parameterized on specific attributes of the target simplification, such as length, paraphrasing, lexical complexity and syntactic complexity are proposed in \cite{martin2019controllable}. 
An attentional encoder-decoder framework with side constraints for politeness is used to control for the level of courtesy in machine translation \cite{sennrich2016controlling}. Textual attributes of sentences such as sentiment, tense, voice, mood and negation are modified by incorporating conditioning information into neural encoder-decoder models \cite{logeswaran2018content}. Generating emotional responses in neural conversational systems is achieved by feeding the emotion category embedding to a sequence-to-sequence decoder \cite{zhou2018emotional}, \cite{asghar2018affective}. Furthermore, informative and on-topic dialogue responses are generated via sentence control functions  \cite{ke2018generating}. Many encoder-decoder models treat the entire dialogue history as one single sequence \cite{serban2016building}, while others treat each conversational turn as a separate sequence \cite{vinyals2015neural}, \cite{shang2015neural},\cite{duvsek2016context}. A topic aware sequence-to-sequence  model is used to generate on-topic conversational responses \cite{xing2016topic}. Topic words are extracted using Latent Dirichlet Allocation and the decoder produces each word conditioned on both the input message
and the topics through a joint attention mechanism. 
Abstractive summaries are produced relying on a pointer-generator network \cite{see2017get}, a hybrid model which combines pointer networks \cite{vinyals2015pointer} to accurately reproduce source side information with a sequence-to-sequence attentional model to generate new words. 

Synthesizing sentences containing specific keywords is done in 
a sequence-to-sequence  backward and forward \cite{mou2016sequence} model by first generating the sentence fragment to the left of the given keyword (in reverse order and conditioned on that keyword), then encoding the sentence fragment generated so far, followed by decoding the sentence fragment to the right of the keyword conditioned on the already generated first part. An encoder-decoder framework for factoid question answering  \cite{yin2016neural} is able to query and generate answers containing terms retrieved from a knowledge base. 
Constrained modification of factual Wikipedia sentences according to given claims is performed via a two-encoder sequence-to-sequence
model with copy attention \cite{shah2020automatic}. Style transfer between scientific paper titles and newspaper titles is approached using multiple decoders for each style, or by passing encoded representations combined with style embeddings to a single decoder  \cite{fu2018style}. Generating memorable headlines constrained on stylistic attributes such as humour, romance and clickbait is performed in a sequence-to-sequence Transformer-based model for stylistic headline
generation \cite{jin2020hooks}. A denoising autoencoding approach is adopted for style transfer which replces  learning disentangled latent representations with back-translation \cite{lample2018multiple}. Neural machine translation models with attention are used for style transfer via backtranslation in the absence of parallel aligned data \cite{zhang2018style}  Neural paraphrase generation is performed through a syntactically constrained encoder-decoder model \cite{iyyer2018adversarial} or via a sequence-to-sequence model with pivoting over multiple sentences from multiple languages \cite{mallinson2017paraphrasing}. Neural poetry translation with a sequence-to-sequence model is proposed in \cite{ghazvininejad2018neural}.
\subsection{GAN Architectures}
\label{gan_nn}

\subsubsection{GAN Models for Generic / Free-Text Generation}

Generative Adversarial Networks (GANs) \cite{goodfellow2014generative} train generative models through an adversarial process which consists of two competing models trained simultaneously: a generative model $G$ that captures the data distribution and whose objective is to generate fake data which is indistinguishable from real data, and a discriminative model $D$ which estimates the probability that a sample came from the training
data rather from the generator $G$. The generator $G(z, \theta_G)$ learns the distribution $p_g$ over real data $\text{x}$ by mapping function the prior noise distribution $p_z(z)$ to real data, while the discriminator $D(x, \theta_D)$ outputs a single scalar value representing the probability that a sample came from the training data instead of $p_g$. The optimization objective for the two-player mini-max game between the $G$ and $D$ can be formally expressed as:

\begin{equation}
\begin{split}
   \min_G \max_D V(D, G) &= \mathbb{E}_{\text{x} \sim p_{\text{data}}(\text{x})} [\log D(\text{x})] \\ &+ \mathbb{E}_{\text{z} \sim p_{z}(\text{z})} [\log (1-D(G(\text{z}))]
\end{split}
\label{gan_objective}
\end{equation}

The GAN framework relies on an implicit probability model, as opposed to incorporating an explicit formulation of the probability density. Unlike autoregressive modeling in which exposure bias is a common issue, in GANs this issue is avoided by sampling synthetic examples at training time from the generator and providing them as input to the discriminator for comparison with real sentences (i.e. sentence-level comparison instead of word-level comparison). The latent representations extracted from real data are distributed according to the specified prior $p_z(z)$ (for eg. Gaussian or uniform). Gradients are backpropagated from the discriminator $D$ through the generated samples
to the generator $G$; note that this is possible only when the generated samples are differentiable w.r.t the $\theta_G$ generator parameters. When the discriminator is trained to optimality before each generator parameter update, the GAN adversarial game is equivalent to minimizing the Jenson-Shannon
divergence between the real data distribution $p_x(.)$ and the synthetic data distribution $p(G(z)), z \sim p_z(.)$ \cite{arjovsky2017towards}. Nevertheless, in such cases gradients can vanish as the discriminator saturates, learns to reject all samples and gives meaningless gradients to the generator. In practice, the second term in Equation (\ref{gan_objective}) is replaced with $\mathbb{E}_{\text{z} \sim p_{z}(\text{z})} [\log (D(G(\text{z}))]$ which helps circumvent the vanishing gradient problem to a certain extent \cite{goodfellow2014generative}.

GAN-based models can be either continuous or discrete, depending on whether the model learns the probability distribution over a sequence of tokens or each individual token $p(x_t|x_{<t})$ at a time. While in the former case it is possible to directly backpropagate from the discriminator into the
generator, in the latter case the generator output is non-differentiable and backpropagation becomes challenging since gradients cannot be passed through the discrete output words of the generator. Therefore, the difficulty of applying GANs to discrete data is caused by the
discontinuity which prohibits the update of the generator parameters via standard back-propagation. The non-differentiability of discrete word tokens
results in difficult generator optimization. Nevertheless, discrete representations are more interpretable and more computationally efficient than their continuous counterparts \cite{jang2017categorical}. Given that the composition of the generator
and discriminator needs to be fully differentiable, existing GAN-based solutions for dealing with discrete data such as text can be categorized as follows: \textit{i)} reinforcement learning based methods, \textit{ii)} latent space solutions, and \text{iii)} continuous approximations of discrete sampling. We present these below. 

\paragraph{Reinforcement Learning (RL) methods}  Discrete GAN models for text generation which employ RL to train the non-differentiable generator represent the current state as the tokens generated so far and the current action is the next token to generate. A discriminator is used to evaluate the current state and provide rewards to the generator to guide its learning. RL-based approaches for training GANs on discrete sequences perform gradient policy update via REINFORCE \cite{williams1992simple} to bypass the generator differentiation problem. Nevertheless, RL training presents its own challenges difficult to deal with, such as the large action space, reward sparsity, the credit assignment problem and large variance for gradient estimation \cite{maddison2016concrete}, \cite{zhang2017adversarial}. Indeed, RL algorithms applied to
dynamic environments with sparse reward are very unstable and the credit assignment problem through discrete computation makes it difficult to pass gradient information to the generator \cite{che2017maximum}. SeqGAN \cite{yu2017seqgan} employs Monte Carlo policy gradient to overcome the differentiation difficulty for discrete data generation, RankGAN \cite{lin2017adversarial} trains the discriminator in a learning to rank setting and evaluates the quality of the generated samples through their relative ranking scores, while LeakGAN \cite{guo2018long} overcomes reward sparsity in a hierarchical RL setting by allowing the discriminator to leak its own high-level features to the generator. Monte-Carlo(MC) rollouts are commonly used to provide ample feedback to the generator at every timestep and circumvent the credit assignment problem. StepGAN \cite{tuan2019improving} proposes a more computationally efficient approach in which the discriminator issues rewards without computing the entire search tree. Pre-training the generator with a negative log-likelihood objective is also commonly used to reduce the large action space and avoid reward sparsity. Nevertheless, RL based methods with pre-training tend to be computationally expensive and more inefficient than solutions based on latent or continuous approximations \cite{haidar2019textkd}. In addition, GAN practical limitations include mode collapse \cite{metz2016unrolled} which occurs when the generator produces same representation for multiple latent representations, and vanishing gradients \cite{arjovsky2017towards} when the
discriminator is close to its local optimum and the generator’s contribution to the learning signal is insignificant; the GAN objective in Equation \ref{gan_objective} thus becomes a weak learning
signal. 

\paragraph{Latent space solutions} These methods extract latent representations of the discrete input data by means of autoencoding and apply smooth transformations to learn the data manifold. Adversarially regularized autoencoders  \cite{zhao2018adversarially} map discrete inputs to an adversarially regularized continuous latent space. TextKD-GAN \cite{haidar2019textkd} uses knowledge distillation on sentence representations learnt by an autoencoder to train the generator to produce similar continuous representations. 

\paragraph{Continuous approximations}  Generating sequences of discrete elements by sampling from a multinomial distribution on discrete objects is not differentiable
with respect to the distribution parameters and is the main limitation why GANs cannot not be directly applied to text generation. Nevertheless, approximating the multinomial distribution with the continuous Gumbel-softmax distribution  \cite{jang2017categorical}, a continuous distribution over the simplex
that can approximate samples from a categorical distribution, is differentiable and allows to backpropagate through samples \cite{kusner2016gans}. In addition, the non-differentiable argmax operator is approximated at learning time with the soft-argmax \cite{zhang2016generating} operator, a continuous differentiable function \cite{zhang2017adversarial}. Other gradient estimators for training neural networks with discrete units include  straight-through estimators \cite{bengio2013estimating}, \cite{raiko2014techniques} and Concrete relaxations \cite{maddison2016concrete}.  

Many GAN models have been proposed to address the task of text generation. Pre-training \cite{yu2017seqgan}, \cite{li2017adversarial}, \cite{yang2018improving} or joint training \cite{lamb2016professor}, \cite{che2017maximum} of the generator and discriminator with a
supervised maximum-likelihood loss is commonly employed before the start of adversarial training, aiming to reduce training instability and guide the generator towards promising improvement directions. Alternatively, in cases when the generator acts randomly, all samples produced by the generator will be easily recognized as fake by the discriminator and consequently a low reward will be assigned to any generator action, resulting in an ineffective training procedure. TextGAN \cite{zhang2017adversarial} leverages a GAN framework (consisting of a LSTM generator and a CNN discriminator) which forces empirical distributions of real and synthetic sentences to have
matched moments in latent-feature space. Instead of optimizing for the GAN objective, MaliGAN \cite{che2017maximum} uses a normalized maximum likelihood target optimized via importance sampling. 
RelGAN \cite{nie2018relgan} uses a relational memory based generator
to model long distance dependencies, along with embedded representations in the discriminator to provide more informative signal to the generator. 

Nevertheless, there are also models using purely adversarial training techniques. ScratchGAN \cite{d2019training} attempts to train language GANs from scratch, i.e. without maximum likelihood pre-training; the model generates realistic looking samples by heavily relying on engineering tricks such large batch sizes for variance reduction, dense rewards provided by a recurrent discriminator at each step for each generated token, and discriminator regularization. A curriculum learning strategy is proposed to generate sequences of increasing length also starting from scratch \cite{press2017language}. Wasserstein GANs-GP  \cite{gulrajani2017improved} 
train a character-level language model in which the discriminator distinguishes between one-hot representations of real text and the probabilistic (softmax) output  vector from the generator; for more stable training the norm of the gradient from the discriminator is penalized. Boundary-seeking GANs \cite{hjelm2018boundary} compute importance weights for the generated samples based on their estimated difference from the generator to use as policy gradients for the generator.    

To overcome reward sparsity, self-adversarial learning \cite{zhou2020self} provides dense rewards to the generator by comparing text quality between
pairs of generated samples similar  (unlike standard GANs which compare fake and real texts). The generator is rewarded by the pairwise comparative discriminator whenever the current generated sentence is better than previously generated samples, similar to a self-play / self-improvement scenario.
In FM-GAN \cite{chen2018adversarial}, latent feature distributions of real and synthetic sentences are minimized by the generator when synthesizing realistic text, and maximized to delineate the dissimilarity of the feature distributions by the discriminator. 


In spite of these attempts at overcoming limitations of GANs for language generation, adversarial learning hurts performance \cite{semeniuta2018accurate}, \cite{garbacea2019judge}. GAN-based models are frequently unstable during training (even less stable than regular language models), extremely sensitive to random initialization and the choice of hyperparameters  \cite{salimans2016improved}, and the error signal provided by the discriminator can be insufficient to train
the generator to produce fluent language \cite{yang2018unsupervised}. In addition, training GANs using gradient-based methods is inherently difficult due to training instability and frequently gradient based optimization fails to converge \cite{salimans2016improved}, \cite{gulrajani2017improved}. In turn, samples produced by GANs are at their best comparable to or even worse in quality than samples produced by  properly tuned conventional language models; the latter are frequently reported in the literature to yield better results than many GAN-based systems \cite{semeniuta2018accurate}. The extent to which GANs generalize from the training data as opposed to memorizing training examples is still an open question \cite{nagarajantheoretical}. Therefore, the benefits of using GANs for language generation are rather unclear, and GAN-based models seem to not benefit much from the maximum likelihood pre-training approach combined with small amounts of adversarial fine-tuning. This in turn suggests that best performing GAN models tend to stay close to the maximum-likelihood training solution \cite{caccia2018language}. Nevertheless, more recent results indicate that relying on pure adversarial training and avoiding the maximum likelihood pre-training step in the GAN training procedure achieves comparable results to maximum likelihood models for unsupervised unconditional word-level text generation \cite{d2019training}.


\subsubsection{GAN Models for Conditional Text Generation}

Conditional GANs \cite{mirza2014conditional} are constructed by feeding the data we wish to condition on $y$ (for eg., class labels or auxiliary data from other modalities) to both the generator and the discriminator as an additional input layer. In the generator $G$ the prior input noise $p_z(z)$ and $y$ are combined in the hidden joint representation, while in the discriminator $D$ the data $x$ and the conditioning information $y$ are specified as different inputs to a discriminative function. The conditional GAN objective function is formulated as:

\begin{equation}
\begin{split}
   \min_G \max_D V(D, G) &= \mathbb{E}_{x \sim p_{\text{data}}(\text{x})} [\log D(\text{x}|\text{y})] \\ &+ \mathbb{E}_{\text{z} \sim p_{z}(\text{z})} [\log (1-D(G(\text{z}|\text{y}))]
\end{split}
\end{equation}

Diverse text generation is encouraged by having a language-model based
discriminator reward the generator based on the novelty of text produced \cite{xu2018dp}. Unlike classifier-based discriminators which in cases when classification accuracy saturates no longer distinguish between relative degrees of novelty, the cross-entropy of the language model does not saturate and is discriminative between repetitive text and novel and fluent text. Generic and uninformative responses are a common problem in dialogue systems. A variational mutual information objective is employed to encourage natural conversations with diverse and unpredictable responses \cite{zhang2018generating}. Adversarial training for open-domain dialogue generation in a reinforcement learning setting is proposed to generate the next response given the dialogue utterance history \cite{li2017adversarial}. To alleviate the credit assignment problem, rewards for each action (word) selection step in  partially decoded sequences are assigned by either using Monte
Carlo search, or by training a discriminator to provide a reward to a partial utterance; nevertheless, computing such a reward is time-consuming. 

MaskGAN \cite{fedus2018maskgan} adopts a fill-in-the-blank approach to text generation and masks contiguous blocks of words in a sentence; an actor-critic conditional GAN 
fills in missing text conditioned on the surrounding context. 
Conditional GANs are used to generate image \cite{dai2017towards} and video \cite{yang2018video} captions. Autoregressive and adversarial models are combined for neural outline generation \cite{subramanian2018towards}. Conditional GANs for neural machine translation with a sentence-level BLEU reinforced objective are proposed in \cite{yang2018improving}. Generating poems from images is accomplished by extracting coupled visual-poetic embeddings and feeding them to a recurrent neural network for poem generation in an adversarial training framework with multiple discriminators via policy
gradient \cite{liu2018beyond}. 

\subsubsection{GAN Models for Constrained Text Generation}

BFGAN \cite{liu2019bfgan} is the first GAN-based model proposed for lexically constrained sentence generation. The model architecture employs two generators, namely a forward generator and a backward generator, as well as a discriminator that guides their joint training and learns to distinguish human-written sentences from machine-generated lexically constrained sentences. The model is used to generate user reviews for Amazon products and conversational responses with lexical constraints. GAN-based stylistic headline generation is proposed in \cite{shu2018deep}.  
\subsection{VAE Architectures}
\label{vae_nn}

\subsubsection{VAE Models for Generic / Free-Text Generation}

The variational autoencoder (VAE) \cite{kingma2013auto}, \cite{rezende2014stochastic}, \cite{doersch2016tutorial}, \cite{kingma2019introduction} is a generative model which integrates stochastic latent variables into the conventional auto-encoder architecture. VAE-based generative models aim to produce realistic samples by feeding noise vectors through the decoder. Given observed variable $\text{x}$, the VAE framework assumes that $\text{x}$ is generated from latent variable $\text{z}$ and models their joint probability as follows:

\begin{equation}
    p_{\theta}(\text{x},\text{z})= p_{\theta}(\text{x}|\text{z})p_{\theta}(\text{z})
\end{equation}

The model is parameterized by $\theta$ and $p_{\theta}(\text{z})$ represents the prior, which is typically chosen to be a simple Gaussian distribution. VAE learns the conditional probability distribution $p_{\theta}(x|z)$ which models the generation procedure of the observed data $\text{x}$ given latent variable $\text{z}$. However, this distribution over latent variables is intractable and VAEs derive an analytic approximation  in the form of recognition model $q_{\phi}(\text{z}|\text{x})$ which estimates latent variable $\text{z}$ for a particular observation $\text{x}$. Probability distributions $p$ and $q$ are parameterized by neural network parameters $\theta$ and $\phi$ (variational parameters), and are learnt by maximizing the variational lower
bound on the marginal log likelihood of data:

\begin{equation}
\begin{split}
    \log p_{\theta} (\text{x}) \ge \mathbb{E}_{\text{z} \sim q_{\phi}(\text{z}|\text{x})} [\log p_{\theta}(\text{x}|\text{z})] \\- \text{KL}(q_{\phi}(\text{z}|\text{x})||p(\text{z})) 
\end{split}
\end{equation}

The KL term ensures distributions estimated by the recognition model $q(\text{z}|\text{x})$ do not diverge from the prior probability distribution $p(\text{z})$ imposed over the latent variables. 
The reparameterization trick is used to train the model with backpropagation and optimize the parameters with gradient descent. According to the reparameterization trick, the Gaussian latent variables $\text{z}$ are reparameterized by
the differentiable functions w.r.t. $\phi$ and are expressed in deterministic form as $\text{z} = g_{\phi}(\epsilon, \text{x})$ with mean $\mu$ and variance $\sigma^2$, where $\epsilon \sim \mathcal{N}(0,1)$ is an independent
Gaussian noise variable. To this end, instead of generating $z$ from $q_{\phi}(\text{z}|\text{x})$, $\text{z}$ is obtained from 
$z=\mu_{\phi}(x) + {\sigma}_{\phi}(x) \circ \epsilon$, allowing gradients to backpropagate through $\phi$. VAEs are considered a regularized version of the standard autoencoder, where the latent variable $\text{z}$ captures the variations $\epsilon$ in the observed variable $\text{x}$. 

While VAEs achieve strong performance in continuous domains, for eg. image modeling \cite{bachman2016architecture}, \cite{gulrajani2016pixelvae}, using VAEs on discrete text sequences is more challenging due to optimization issues. Parameterizing
conditional likelihoods with powerful function approximators such as neural networks makes posterior inference intractable and introduces points of non-differentiability which complicate backpropagation \cite{kim2018tutorial}. In particular, the collapse of the KL divergence term in the latent loss to zero leads to the model behaving like a regular language model and completely ignoring the latent representations \cite{bowman2015generating}, \cite{pelsmaeker2019effective}. This posterior collapse issue occurs in particular when learning VAEs with an auto-regressive decoder, and in such cases the model generates repetitive and uninteresting samples \cite{semeniuta2017hybrid} and behaves like a regular language model \cite{pelsmaeker2019effective}. In addition, the assumption that the variational posterior is Gaussian introduces an approximation gap with respect to the true posterior \cite{cremer2018inference}.
Solutions proposed in the literature aim to force the decoder to incorporate the information from the latent vectors by imposing structured sparsity on the latents \cite{yeung2016epitomic}, batch normalization and deterministic warm-up to gradually turn on the KL-term \cite{sonderby2016ladder}, as well as input dropout \cite{bowman2015generating} and adding auxiliary reconstruction
terms computed from the activations of the last decoder layer \cite{semeniuta2017hybrid}. Nevertheless, training deep latent variable models for discrete structures is still an open research problem \cite{zhao2018adversarially}. 

An RNN-based variational autoencoder generative model is used to generate natural language sentences from a latent continuous space \cite{bowman2015generating}. To this end, distributed latent representations encode the full content of sentences and allow to explicitly incorporate and vary textual attributes such as style, topic,
and high-level syntactic features. Nevertheless, the authors report the negative result that VAEs with LSTM decoders perform worse than LSTM language models; this is attributed to LSTM decoders ignoring the conditioning information from the encoder. In follow-up work, VAEs outperform language models when the decoder architecture is changed with a dilated CNN, demonstrating the trade-off between the effectiveness of encoding information and the contextual capacity of the decoder \cite{yang2017improved}. For generating longer texts, a VAE framework based on a convolutional encoder and
a decoder which combines  deconvolutional and RNN
layers is used in \cite{semeniuta2017hybrid}. The variational RNN \cite{chung2015recurrent} incorporates random latent variables into the hidden state of a recurrent neural network and is designed to model variability in highly structured sequential data such as speech. 

Finally, it is important that the posterior distribution over latent variables
appropriately covers the latent space \cite{bowman2015generating}. When mapping sentences to latent representations there are many regions in the latent space which do not necessarily map or decode to realistic-looking sentences, therefore it is not enough to only cover a small region of the latent space corresponding to a manifold embedding  \cite{zhang2017adversarial}. VAEs for text generation are also difficult to train when combined with powerful autoregressive decoders --  ``posterior collapse'' causes the model to rely entirely on the decoder and ignore latent variables. Latent space expanded VAE disperses sentences into the latent space based on their similarity to avoid mode collapse \cite{song2019latent}.

\subsubsection{VAE Models for Conditional Text Generation}

In the standard VAE model it is difficult to control textual features directly since the latent code is assumed to be Gaussian distributed; this makes it impossible to distinguish which
part of code controls the structure and which part
controls the semantics \cite{li2019topic}.  

A document-level language model based on the VAE architecture is introduced in \cite{miao2016neural} for the answer selection problem. The model represents texts as bags of words and extracts a continuous semantic latent
variable for each document which is then passed to decoder which generates either generic or conditional sentence reconstructions. Auto-encoding sentence compression (both extractive and abstractive) is modeled in the VAE framework by first drawing a compact summary sentence
from a latent background language model, and then drawing the observed sentence conditioned on the latent summary \cite{miao2016language}. Variational neural machine translation \cite{zhang2016variational} incorporates a continuous latent variable to learn the conditional distribution
of a target sentence given a source sentence and learn the underlying
semantics of sentence pairs. In follow-up work, the variational recurrent neural machine translation model introduces a sequence of continuous random latent variables $z=\{z_1, z_2, \ldots, z_N\}$ to
capture the underlying semantics of sentence pairs and model the high variability in structured data \cite{su2018variational}. Conditional VAEs that condition on observed images are proposed for image caption generation
\cite{pu2016variational}. A conditional VAE is introduced for the task of poetry generation \cite{liu2019rhetorically}, conditioning on aesthetical aspects of the generated poem such as the use
of metaphor and personification.

\subsubsection{VAE Models for Constrained Text Generation}


Semi-supervised VAEs that operate on both continuous and discrete latent variables are used for labeled sequence transduction -- given an input sequence and a set of labels, the model changes the input sequence to reflect attributes of the given labels \cite{zhou2017multi}. VAEs are enhanced with attribute discriminators that help the model learn disentangled latent representations of semantic structures \cite{hu2017toward}. In addition, this also helps enhance interpretability in the latent space where each attribute is focusing solely on just one aspect of the generated samples; authors control for the sentiment and tense of the generated sentences. Implicit latent features in VAEs are extracted following a sample-based approach which aligns the posterior to the prior distribution \cite{fang2019implicit}. Topic guided variational autoencoder \cite{wang2019topic} is used for text generation on a specific topic of interest. Unlike the VAE which specifies a simple Gaussian prior for the latent code, the model specifies the prior as a Gaussian mixture model parameterized by
a neural topic module. Style transfer for tasks such as sentiment modification, word substitution and
word ordering is achieved using a VAE model that separates content from stylistic properties of text \cite{shen2017style}. To this end, the VAE encoder takes as input a sentence and its original style indicator and maps it to
a style-independent content representation; this representation is then passed to a style-dependent decoder for
generation. Learning disentangled representations for style transfer are also proposed in \cite{balasubramanian2020polarized}, \cite{john2019disentangled}. Paraphrase generation is performed through a VAE module with two latent variables
designed to capture semantics and syntax \cite{chen2019controllable}, \cite{bao2019generating}.

\subsection{Memory-based Architectures}
\label{memory_nn}


Although RNNs and LSTMs are trained to predict the next token in a sequence, their memory is small and used mainly to store information about
local context, which does not allow these models to accurately recall facts from the past. Indeed, recurrent neural network memory degrades with time \cite{khandelwal2018sharp}. Parametric neural networks implicitly encapsulate memory in their weights, nevertheless this hurts their ability to generalize across complex linguistic tasks \cite{nematzadehmemory}. Attempts to capture non-local dependencies in language models aim to enhance their ability to adapt to a changing environment and dynamically update the word probabilities based on the long-term context.  
Improving neural language models with external storage units is done by means of introducing an external memory component in the form of a \textit{soft attention mechanism} \cite{bahdanau2014neural}, \cite{luong2015effective}, \cite{daniluk2017frustratingly} which allows them to focus on specific parts of the input,  an \textit{explicit memory block} which implicitly captures  dependencies for word prediction \cite{tran2016recurrent}, or \textit{cache model} \cite{grave2016improving} which can be added on top of a pre-trained language model. Shared memory models are reported to further improve attention based neural models \cite{munkhdalai2017neural}.

Integrated LSTM networks are proposed to alleviate the practical engineering requirements of LSTMs by relying on external memory units to enhance the memory capacity of neural networks. Neural Turing Machines \cite{graves2014neural} extend the memory resources of RNNs by coupling them with an addressable external memory bank that can be read from and written to (i.e. random access memory with read and write operations). C-LSTMs \cite{zhou2015c} combine CNN with LSTM networks to learn high-level sentence representations that capture both local features of phrases and global and temporal sentence semantics. In the context of question answering, the use of a long-term memory acting similar to a dynamic knowledge base which can be read from and written to is proposed in memory networks \cite{weston2014memory}. Nevertheless, the discrete model is difficult to train via backpropagation and requires supervision at each layer of the network. The memory network architecture is further extended to operate without supervision in a continuous space \cite{sukhbaatar2015end}.
Single-layer LSTM networks enhanced with an unbounded differentiable memory, yield comparable performance to deep RNNs in sentence transduction tasks such as machine translation \cite{grefenstette2015learning}. Memory based architectures incorporating stacked layers of memories for storing and accessing intermediate representations in sequence-to-sequence learning are proposed in \cite{meng2015deep}. Dynamic memory networks \cite{kumar2016ask} are used to generate relevant answers in question answering by means of episodic memories  reasoned over in a hierarchical recurrent sequence model. 

Memory architectures for recurrent neural network language models are compared in \cite{yogatama2018memory}. Stack-based memory access which dynamically stores and retrieves contextual
information with a stack is shown to outperform sequential access which fails at capturing long term dependencies or random memory access in which the learner needs to infer dependencies from the data in the absence of any structural biases. Instead of having a monolithic model to fit all training examples, a few-shot meta-learning scenario in which 
multiple task-specific models covering groups of similar examples is proposed in \cite{huang2018natural}. 

While the on-going trend in language modeling is to learn contextual representations from ever larger datasets, alternative methods which are sample efficient and leverage smaller amounts of data represents the next research frontier for deep learning models. $k$NN-LMs \cite{khandelwal2019generalization} is a general framework which allows to augment any pre-trained language model by means of linearly interpolating its next
word distribution with a k-nearest neighbors search. The approach helps memorize long-tail patterns (e.g., factual knowledge and rare $n$-grams) explicitly by drawing nearest neighbours from any text collection in the pre-trained embedding space  rather than modeling these rare patterns implicitly in the model parameters.

An additional memory component is used to store external simplification rules from a paraphrase database in neural text simplification in combination with the multi-layer and multi-head attention Transformer architecture  \cite{zhao2018integrating}; the additional memory is used to recognize the context and output of each simplification rule. Neural semantic encoders \cite{munkhdalai2017neural} augment neural network models with an evolving memory of the input sequence for natural language understanding tasks including natural language inference, question answering, sentence classification, sentiment analysis and machine translation.  Relational memory \cite{santoro2018relational} adds interactions between memory units via attention and is designed to enhance reasoning abilities of neural networks across sequential information. An external factual memory component is incorporated into a neural pre-trained language model for question answering \cite{verga2020facts}.
Finally, memory networks are used to generate scientific articles with constraints on entities and human-written paper titles \cite{wang2019paperrobot}.


\subsection{Reinforcement Learning (RL) Architectures}
\label{rl_nn}
\subsubsection{RL Models for Generic / Free-Text Generation}

Reinforcement learning is used in the context of natural language generation to directly optimize non-differentiable reward functions and evaluation metrics.  To this end, policy gradient methods such as REINFORCE \cite{williams1992simple} are used to alleviate current issues in training  generative models for text generation, namely exposure bias and loss functions which do not operate at the sequence level. In the RL framework the generative model is seen as an agent with parameters that define a policy and which interacts with an external environment by taking actions, receives a reward once it reaches the end of a sequence and  updates its internal state consequently. While any user-defined reward function can be employed for training, most frequently optimized metrics with RL are BLEU for machine translation and image captioning \cite{ranzato2015sequence}, \cite{wu2016google}, ROUGE for text summarization \cite{ranzato2015sequence}, \cite{paulus2017deep}.  Nevertheless, policy gradient algorithms present large variance and generally struggle in settings with large action spaces such as natural language generation. In addition, the improvement in the optimized metrics is not always reflected in human evaluations \cite{wu2016google}. In the context of machine translation in particular, reinforcement learning methods do not optimize the expected reward and take very long time to converge  \cite{choshen2019weaknesses}.

\subsubsection{RL Models for Conditional Text Generation}

Deep reinforcement learning is used to model the future reward in neural conversational systems and reward  responses that are informative, coherent and simple \cite{li2016deep}. Algorithms such as DQN \cite{zhao2016towards}, \cite{li2017end}, \cite{peng2018deep}, \cite{cuayahuitl2016deep}, policy-gradient \cite{liu2017end} and actor-critic \cite{peng2018adversarial}, \cite{liu2017iterative} have been widely used for single-domain or multi-domain dialogue generation for movie-ticket bookings and restaurant search.   

Inverse reinforcement learning produces more dense reward signals and generates texts with higher diversity \cite{shi2018toward}. Inverse reinforcement learning has been applied in paraphrase generation \cite{li2018paraphrase} and in open-domain dialogue systems   \cite{li2019dialogue} for modeling the reward function. Reward functions are learnt from human preferences and further optimized with reinforcement learning for tuning language models \cite{ziegler2019fine} and text summarization \cite{bohm2019better}. Conditional RNNs are trained via REINFORCE to directly optimize for test time evaluation metrics such as BLEU for machine translation and image captioning, and ROUGE for text summarization tasks \cite{ranzato2015sequence}. 
 
Sequence-to-sequence models can be further improved with reinforcement learning training to alleviate exposure bias and improve generalization \cite{keneshloo2019deep}.
Neural machine translation is framed as a stochastic reinforcement learning policy with translation adequacy rewards  \cite{kong2019neural}. Generating polite dialogue responses is encouraged by rewarding polite utterances with positive reward, and rude
ones discouraged with negative reward \cite{niu2018polite}. Internal rewards such as ease of answering, semantic coherence, emotional intelligence, as well as external rewards based on human feedback are incorporated through reinforcement learning in an encoder-decoder framework for dialogue response generation focused on movie and restaurant reviews \cite{srinivasan2019natural}. Dialogue responses are generated by conditioning on discrete attributes such as sentiment, emotion, speaker id, speaker personality and user features   when framing dialogue attribute selection as a reinforcement learning problem \cite{sankar2019deep}. Abstractive headline generation is performed in a reinforcement learning setting which maximizes the
sensationalism score as the reward for the reinforcement learner \cite{xu2019clickbait}. 


Hierarchical models decompose the learning problem into a sequence of sub-problems and are a natural fit to language given its  hierarchical structure. These models first decompose natural language into a sequence of utterances, and then decompose each utterance into a sequence of words. Hierarchical reinforcement learning is employed in open-domain dialogue generation to optimize for human-centered metrics of conversation quality and prevent the generation of inappropriate, biased or offensive language \cite{saleh2019hierarchical}. Hierarchical sequence-to-sequence dialogue models   are employed to learn reward functions from human interactions \cite{jaques2019way}. Task-oriented dialogue systems \cite{peng2017composite}, \cite{budzianowski2017sub}, \cite{tang2018subgoal}, \cite{zhang2018multimodal} learn hierarchical dialogue policies  by diving a complex goal-oriented task into
a set of simpler subgoals with distinct reward functions; these methods have been applied to various dialogue tasks such as travel planning or task-oriented visual dialogue.  



\subsubsection{RL Models for Constrained Text Generation}

Non-monotonic constrained text generation is framed as part of an immitation learning framework (learning a generation policy that mimics the actions of an oracle generation policy) in which a token is first generated in an arbitrary position in the sentence, and the model recursively generates a binary tree of words to its left and right \cite{welleck2019non}. 

Extractive and abstractive sentences are mixed in a hierarchical reinforcement learning framework for text summarization in which a copy-or-rewrite mechanism allows to switch between copying a sentence and rewriting a sentence \cite{xiao2020copy}. Policy gradient methods that optimize non-differentiable evaluation metrics (for eg., ROUGE) are used for extractive summarization in a contextual bandit setting \cite{dong2018banditsum} 
or in a sentence ranking setting 
\cite{narayan2018ranking}, as well as for abstractive summarization in a hierarchical setting \cite{chen2018fast}. Saliency and logical entailment rewards for abstractive summarization are simultaneously optimized by means of reinforce-based policy gradient \cite{pasunuru2018multi}. 
A hybrid learning objective which combines standard supervised word prediction using maximum likelihood with reinforcement learning policy gradient is used for abstractive summarization 
\cite{paulus2018deep}, \cite{celikyilmaz2018deep}. To improve the level of abstraction in summary generation, a ROUGE based reward is combined with a novelty metric which counts the fraction of unique $n$-grams
in the summary that are novel in the policy gradient optimization objective  \cite{kryscinski2018improving}. A cycled reinforcement learning approach with unpaired data is proposed for the task of sentiment-to-sentiment translation to generate emotionally charged sentences \cite{xu2018unpaired}.

\subsection{Transfer Learning for NLG}
\label{pretrained_nn}


\subsubsection{Transfer Learning Models for Generic / Free-Text Generation}

Recent advances in natural language generation rely on pre-training a large
generative language model on a large corpus of unsupervised data, followed by fine-tunning the model for specific applications.
The goal of pre-training is to provide models with general purpose knowledge that can be leveraged in many downstream tasks \cite{raffel2019exploring}. Indeed, large-scale language models pre-trained on huge unlabeled datasets have shown unparalleled text generation capabilities substantially outperforming training on supervised datasets from scratch and have considerably advanced the state-of-the-art on many natural language processing problems. These models leverage the Transformer architecture \cite{vaswani2017attention} pre-trained on large amounts of text and optimize for different unsupervised language modeling objectives, showing that transferring many self-attention blocks can often replace task-specific architectures \cite{devlin2018bert}, \cite{radford2019language}. High-capacity  language models pre-trained on large datasets can be an alternative to traditional knowledge bases extracted from text \cite{petroni2019language}. These models can acquire commonsense reasoning capabilities about previously unseen events \cite{sap2019atomic}, infer relations between entities \cite{jiang2019can}, \cite{soares2019matching}, \cite{rosset2020knowledge}, answer factoid questions and commonsense queries \cite{trinh2018simple}, as well as open-domain questions without access to any external context or knowledge source \cite{roberts2020much}. Nevertheless, while these models leverage massive amounts of data and excel at capturing statistical patterns in the datasets, they are sample inefficient and fail to generalize as quickly and robustly as humans \cite{linzen2020can}. 


\subsubsection{Transfer Learning Models for Conditional Text Generation}



Early work \cite{ramachandran2017unsupervised} shows that pretraining improves the generalization of
sequence-to-sequence models. Using unsupervised learning to initialize the weights of both the  encoder and decoder with the pretrained
weights of language models outperforms purely supervised learning baselines for machine translation and abstractive summarization. Moreover, language modeling pre-training is also helpful for difficult text generation tasks such as chit-chat dialog and dialog
based question answering systems \cite{dinan2018wizard}, \cite{wolf2019transfertransfo}. 

Representation models for language successfully adopt a masked language modeling approach similar to denoising auto-encoding \cite{vincent2008extracting}, in which the identities of a subset of input tokens are masked and a neural network is trained to recover the original input. BERT \cite{devlin2018bert} is a multi-layer bidirectional Transformer \cite{vaswani2017attention} encoder used for learning deep contextualized token representations from unlabeled text; the model incorporates left and right context fusion to predict the
masked words. 
Nevertheless, the bidirectional nature of BERT does not allow to use the model as is for text generation purposes
\cite{wang2019bert}.
BERT is used for sequence
generation for text summarization as part of a pre-trained encoder-decoder framework which relies on a BERT-based encoder and a Transformer-based decoder \cite{zhang2019pretraining}. A similar BERT-based encoding approach is adopted for both extractive and abstractive text summarization in \cite{liu2019text}. In parallel, masked
sequence-to-sequence pre-training \cite{song2019mass} proposes a BERT inspired pre-trained objective in an encoder-decoder framework in which the decoder is trained to reconstruct an encoded sentence with randomly masked fragments; the model is applied to generative tasks such as neural machine translation, text summarization and conversational response generation. UniLM \cite{dong2019unified} extends BERT for sequence generation by combining unidirectional,   bidirectional and sequence-to-sequence unsupervised language modeling objectives. Building upon the success of natural language models, a wide range of models are proposed for jointly modeling vision and language tasks, including VisualBERT \cite{li2019visualbert}, ViLBERT \cite{lu2019vilbert}, VideoBERT \cite{sun2019videobert} for image and video captioning, visual question answering and visual commonsense reasoning. 

OpenAI-GPT \cite{radford2018improving}, \cite{radford2019language}, \cite{brown2020language} autoregressive models learn universal representations from massive unlabeled datasets useful for a wide range of language tasks such as text summarization, machine translation, question answering and reading comprehension. These models build upon the left-to-right Transformer \cite{vaswani2017attention} to predict a text sequence word-by-word initially in a semi-supervised fashion \cite{radford2018improving}, by combining unsupervised generative pre-training on a large unlabeled text corpus with supervised discriminative fine-tuning for quick adaptation to a particular task. Later extensions \cite{radford2019language}, \cite{brown2020language} are completely unsupervised and demonstrate the ability to adapt to few-shot and zero-shot settings even without fine-tuning in a multitude of text generation scenarios, including machine translation, text summarization, question answering and news story generation.

Many other   extensions demonstrate that Transformers can be used for generative
tasks. Transformer Memory Networks \cite{dinan2018wizard} combine the Transformer  architecture with memory networks \cite{sukhbaatar2015end} in the context of dialogue agents that store encyclopedic knowledge in large memory systems and carry engaging  open-domain conversations. Transformer-XL \cite{dai2019transformer} captures longer term dependencies by adding recurrence into the deep self-attention network. A BERT initialized Transformer model  is proposed for text simplification \cite{jiang2020neural}. BART \cite{lewis2019bart} relies on the Tranformer architecture to train a sequence-to-sequence denoising autoencoder for tasks such as abstractive dialogue generation, question answering, machine translation and text summarization. Turing-NLG  \cite{turingnlg2020} is a Transformer based generative language model useful in text summarization and question answering. More efficient versions which improve memory and time constraints for long-term
structure  generation are proposed by Sparse Transformers \cite{child2019generating}, Reformer \cite{kitaev2019reformer}, Universal Transformers \cite{dehghani2018universal}, Compressed Transformer \cite{rae2019compressive}, Evolved Transformer \cite{so2019evolved}, Megatron-LM \cite{shoeybi2019megatron}, Big Bird \cite{bigbird}, \cite{rae2020transformers}. Wikipedia articles are generated using a decoder-only sequence transduction model  by conditioning on the article title \cite{liu2018generating}. Furthermore, Transformer language models are found to outperform sequence-to-sequence models for neural document summarization \cite{subramanian2019extractive}. Conditional Transformer is used to control for attributes of the generated text such as style and content \cite{keskar2019ctrl}. Similarly, the pre-trained Transformer is combined with attribute classifiers to control attributes
of the generated language \cite{dathathri2019plug}. Lexical and syntactic constraints are added to the Transformer architecture to control the type and level of text simplification  \cite{mallinson2019controllable}. A hierarchical Transformer encoder is used for multi-document summarization \cite{liu2019hierarchical}. MARGE \cite{lewis2020pre} proposes a self-supervised alternative to the masked language modeling objective by reconstructing the target text conditioned on retrieved related documents, and is used for machine translation, text summarization, question answering and paraphrasing. Any natural language processing task can be formulated as a ``text-to-text'' generation problem, feeding text as input and producing new text as output in T5 \cite{raffel2019exploring}. In addition, adversarial pre-training on top of Transformer-based language models by applying perturbations in
the embedding space improves robustness and generalization \cite{liu2020adversarial},
\cite{wang2019improving}. 

Flexible sequence generation in arbitrary orders with dynamic length changes and refinement through insertion and deletion operations is introduced in Levenstein Transformer \cite{gu2019levenshtein}. Similarly, generating sequences in the absence of a predefined generation and through iterative refinement in multiple passes is proposed in \cite{emelianenko2019sequence}, \cite{ford2018importance}. Furthermore, neural network-based pre-trained language models can act like universal and general-purpose decoders for generative tasks \cite{raffel2019exploring} and can be steered to recover arbitrary sentences \cite{subramani2019can}. The main components of Transformer’s attention and the evolution of representations learnt across layers are analyzed in \cite{tsai2019transformer},  \cite{voita2019bottom}, \cite{kaplan2020scaling}, \cite{talmor2019olmpics}, \cite{yogatama2019learning}. 

\subsubsection{Transfer Learning Models for Constrained Text Generation}

Recent progress in Transformer-based language models has led to generative models that learn powerful distributions and produce high quality samples. While these large scale language models display promising text generation capabilities, it is desirable to allow the user to control different aspects of the generated text and include user-defined key phrases in the generated output. 

Soft-constrained text generation by integrating external knowledge into a neural conversational model is achieved by encoding the conversation history and relevant external text excerpts, and passing them both to a Transformer-based response generator  \cite{qin2019conversing}.  Counterfactual story generation  does minimal revisions to an existing story constrained on a given intervening counterfactual event. OpenAI-GPT2 \cite{radford2019language} pre-trained model is used to  re-write a story through counterfactual reasoning and make the narrative 
consistent with the imposed constraints \cite{qin2019counterfactual}. OpenAI-GPT2 \cite{radford2019language} is also used for abstractive summarization in a reinforcement learning setting which trains the summarization agent to maximize coverage and fluency constrained on a given length \cite{laban2020summary}.
Hard-constrained text generation under specified lexical constraints is performed by using a masked language modeling objective \cite{devlin2018bert} which recursively inserts new tokens between existing ones until a sentence is completed
\cite{zhang2020pointer}. Sentence generation is carried in a hierarchical fashion, by first generating high-level words (nouns, verbs, adjectives), using them as pivoting points for iteratively inserting finer granularity details, and finally adding the least informative
words (pronouns and prepositions).


\subsection{Discussion - Neural NLG}
\label{nlg_discussion}

Advances in the field of deep learning have reignited the hopes of having machine models capable to generate realistic and coherent natural language. The field of natural language generation has undergone major changes in recent years, and is currently witnessing impressive developments and an increased surge in interest. The availability of large and diverse datasets, combined with powerful neural models and compute-intensive  infrastructure have led to high-capacity neural
models achieving  widespread success in a multitude of language generation tasks, from machine translation, text summarization, to dialogue generation and creative applications such as story and poetry generation. 

In this section we have presented the latent developments in natural language generation and introduced the models employed for generating texts that fulfill various user goals in a multitude of problem scenarios. Significant performance gains are reported by using larger and larger models, on datasets larger than ever before. Consequently, it is not
clear where the ceiling is when combining pre-training with finetuning approaches \cite{radford2019language}. Nevertheless, conducting comparisons between the various neural approaches to natural language generation is challenging, since it is not always possible to reproduce the reported results, datasets used are not always publicly available, and the impact of many key hyperparameters and training data size presents a significant impact on the final performance \cite{liu2019roberta}. 

Despite the reported recent success, natural language generation remains a difficult problem to model and there is still a large gap to achieving human peformance \cite{turing1950computing}, \cite{linzen2020can}. Generating long and coherent pieces of text that capture long-term
dependencies in the data is particularly challenging. Longer generated texts are frequently incoherent and present grammatical errors, lack in diversity, include  redundant, short and safe phrases, and contradictory arguments. Unsurprisingly, powerful neural models tend to memorize the training data and often fail to generalize and demonstrate that they learn meaningful representations that capture more than just shallow patterns in the data. Moreover, these systems are brittle, sensitive to slight changes in the data distribution and task specification \cite{radford2019language}. The lack of generalization is also directly tied to their inability to perform natural language understanding and inference, both important hallmarks of intelligence. Natural language understanding requires mastery of linguistic structure and the ability
to ground it in the world, and meaning cannot be learnt solely by relying on huge training datasets   \cite{bender2020climbing}. In addition, robust evaluation metrics which can accurately measure the ``goodness'' of the generated language are imperative for quantifying research progress, comparing natural language generation models and pushing forward the state-of-the-art.   


While current generative models display promising free-form text generation abilities with rather little conditioning beyond the input context on the generated output, it is desirable to produce output conditioned or constrained on particular text attributes for the generation of meaningful texts in specific contexts. To this end, modeling and manipulating the stylistic properties of the generated text also reflects how humans  communicate with a specific intent or goal in mind. Conditional and constrained text generation are important research directions for better human-AI interaction which allow users to control the content and style of the generated text. Furthermore, approaches that incorporate external information in the generation process enhance language representations with structured knowledge facts for more general and effective language understanding \cite{sun2019ernie}, \cite{lewis2020retrieval}, \cite{rosset2020knowledge}. 

The current trend nowadays is to train bigger models on ever larger datasets, however large models are not necessarily more robust to adversarial examples \cite{jia2017adversarial} and their behaviour is unpredictable and inconsistent when the test set distribution differs
from the training data distribution \cite{mccoy2019right}. Instead, training these models on diverse datasets has the potential to improve out-of-distribution robustness \cite{hendrycks2020pretrained}. To this end, careful consideration is required when selecting the training data to ensure fair and unbiased language generation, and responsible research and innovation \cite{brundage2016artificial}. For example, relying on uncurated movie scripts or dialog datasets collected online for training models often leads to malicious, aggressive, biased or offensive responses
\cite{blodgett2020language}. Abstractive summarization models tend to generate untruthful information and fake facts when fusing parts of the source document  \cite{cao2018faithful}. Societal biases such as race, gender and age are often encoded in the word embeddings used by the generative models \cite{romanov2019s}. To prevent such problems, an increased focus on the fairness, accountability, and transparency issues of generative systems is essential and imperative . 

Neural generative models, especially large-scale pre-trained models, encode commonsense knowledge and factual and relational information in their latent parameters \cite{petroni2019language}, \cite{roberts2020much}. However, inspecting and interpreting this information is not straightforward  \cite{lei2017interpretable}. Adding interpretability to neural models 
can increase user's acceptance of the models and trust in their ability to make informed decisions \cite{reiter2019natural}. To this end, natural language generation can help with providing human-interpretable explanations for neural generative or discriminative models \cite{forrest2018towards}.




Finally, it is important to consider text generation for low resource languages or  tasks for which large datasets are not readily available \cite{tilk2017low}. There is a huge gap between neural generative models's ability to generalize quickly and robustly in low-resource settings compared to human's ability to learn language from limited exposure to data \cite{linzen2020can}. Performing text generation in few-shot or zero-shot settings in an important step towards having general systems which can perform many tasks \cite{radford2019language}, eventually
without the need to manually create and annotate a training
dataset for each task in particular \cite{brown2020language}. Having competent general-purpose systems which can perform many tasks and which can easily generalize to new domains instead of relying on specialized narrow expert systems is a longstanding dream of artificial intelligence. Recent progress shows that designing task specific architectures can be replaced with large models able to simultaneously accomplish a multitude of natural language generation tasks in diverse domains \cite{domains}. Moreover, the OpenAI GPT2 \cite{radford2019language} model shows that architectures designed specifically for text generation can be straightforwardly used for image generation too \cite{chengenerative}.

We hope to see progress in the future on the research directions outlined for developing robust and fair natural language generation systems.

\section{Evaluation Methods}
\label{evaluation}

While many natural language generation models have been proposed in the literature, a critical question is what objective metrics to use for their evaluation and for meaningful comparison with other models. Choosing the appropriate model is important for obtaining good performance in a specific application, nevertheless the choice of the evaluation metric is equally important for measuring progress and drawing the right conclusions. As we are witnessing considerable progress in the field of natural language generation, evaluation  of the generated text is largely an unsolved problem. Currently, there is no consensus on how NLG systems should be evaluated \cite{van2019best}, \cite{gkatzia2015snapshot}, and the lack of meaningful quantitative evaluation methods to accurately assess the quality of trained models is detrimental to the progress of the field. In the absence of well established evaluation measures, natural language evaluations are carried in a rather ad-hoc manner with a lot of variability across the proposed models and tasks, resulting in misleading performance measures. Subjective evaluations based on visual inspection of the generated samples are often carried, making it difficult to quantify and judge precisely the quality of a generative model \cite{hashimoto2019unifying}. In addition, the evaluation of generative models is a notoriously difficult problem \cite{borji2019pros}. 

The two main approaches to performance evaluations are based on either intrinsic or extrinsic criteria. While intrinsic criteria relate to a system's objective, extrinsic criteria focus to its function and role in relation to the purpose it was designed for \cite{galliers1993evaluating}. In what follows we summarize these  approaches to natural language evaluation, starting with intrinsic evaluation in Section \ref{intrinsic_eval},  continuing with extrinsic evaluation in Section \ref{extrinsic_eval}, and finally summarize these approaches and  main takeaways in Section \ref{eval_discussion}.

\subsection{Intrinsic Evaluation}
\label{intrinsic_eval}
Intrinsic measures of evaluation assess properties of the system or system components in terms of the output produced, and can be further categorized into user like measures (human-based, subjective assessment of quality) or output quality
measures (corpora-based, carried automatically) \cite{belz2014towards}. We provide a detailed overview of intrinsic evaluation metrics below.

\subsubsection{Human Evaluation}

Human evaluation of generative models is a straightforward surrogate of the Turing test in which human judges are asked to assess whether machine-generated samples can be distinguished from real data. Human evaluations measure either holistic properties of the generated text, such as overall quality, or are conducted at a more finer-grained level to measure particular attributes such as fluency, relevance \cite{dathathri2019plug}, adequacy, correctness, informativeness, naturalness, meaning preservation, simplicity, grammaticality, degree of realism \cite{novikova2017we}. Human evaluations are commonly regarded as the gold standard for generative models, however there is a high degree of variation in the way human evaluations are conducted \cite{van2019best}. Furthermore, these evaluations are expensive to carry and it is impossible to thoroughly assess through human evaluations any generative model across the entire quality-diversity spectrum. Typically only a few samples generated by the model are presented to human raters, allowing to measure precision and sample quality, but not recall and diversity. In addition, it is impossible to identify models which simply plagiarize the training set; due to this human evaluations may yield unrealistically optimistic scores \cite{semeniuta2018accurate}. Human crowdsourcing evaluations are proposed to assess generative realism in HYPE \cite{zhou2019hype}. Best practices for carrying human evaluations are summarized in \cite{van2019best}. With the latest advances in natural language generation, it is frequently reported in the literature that human evaluators have difficulty in identifying machine-generated sentences in the domain of short stories \cite{donahue2020enabling} or online product reviews \cite{garbacea2019judge}.

\subsubsection{Automated Evaluation Metrics}

Automatic evaluation is a quicker and cheaper alternative compared to human evaluation. Nevertheless, the use of automated evaluation metrics is dependent upon their correlation with human judgements of quality. To this end, there is a wide variety of factors that influence the correlation of automatic evaluation metrics with human judgements \cite{fomicheva2019taking}, including domain, type of human evaluation employed and its reliability, type of machine learning system assessed, language pair (for machine translation), or correlation metric used which can be unstable and highly sensitive to outliers \cite{mathur2020tangled}. From a machine learning perspective, automated evaluation can be divided into learnable and non-learnable evaluation metrics. While non-learnable evaluation metrics rely on heuristics / manually defined equations to measure the quality of the generated sentences, learnable metrics train machine learning models to immitate human judgements. 

\paragraph{N-gram based metrics}
Metrics measuring word overlaps were originally developed in the machine translation community to estimate surface similarity between the translated texts and a set of ground-truth human-written references in the target language, and are currently adopted for the evaluation of the generated text in a multitude of tasks. These metrics assume the existence of a  human-written set of references which is often not available \cite{xu2016optimizing}, and make strong assumptions regarding its correctness and completeness \cite{novikova2017we}. BLEU \cite{papineni2002bleu}, SentBLEU \cite{lin2004orange}, $\Delta$ BLEU \cite{galley2015deltableu}, NIST \cite{doddington2002automatic}, ROUGE \cite{lin2004looking},  METEOR \cite{banerjee2005meteor}, \cite{lavie2009meteor}, \cite{denkowski2014meteor}, \cite{guo2019meteor}, SERA \cite{cohan2016revisiting}, LEPOR \cite{han2012lepor}, CIDEr \cite{vedantam2015cider}, SPICE \cite{anderson2016spice}, SPIDER \cite{liu2017improved}, SARI \cite{xu2016optimizing}, RIBES \cite{isozaki2010automatic}, MPEDA \cite{zhang2016extract} are commonly used to assess sample quality based on comparisons with a set of human-written references in many natural language generation tasks such as dialogue systems \cite{song2016two}, \cite{tian2017make}, machine translation, text summarization, text simplification, image captioning. Metrics measuring $n$-gram overlap at character level are also proposed in CHRF \cite{popovic2015chrf}, CHRF++ \cite{popovic2017chrf}.  As demonstrated by numerous studies, $n$-gram matching is not adequate for the evaluation of unsupervised language generation models as it fails to capture semantic variation. Indeed, BLEU scores are insufficient in evaluating text generative systems \cite{ehudreiter2020}, 
lack interpretability, do not detect deterioration in sample quality \cite{van2019best} and overall are not representative of the quality of a model  \cite{semeniuta2018accurate}.
Multiple studies also show that $n$-gram based metrics correlate poorly with human judgements at the instance-level \cite{novikova2017we}, \cite{stent2005evaluating}, \cite{specia2010machine}, \cite{wu2016google}, \cite{liu2016not}, fail to account for semantic similarity \cite{chaganty2018price}, do not capture diversity \cite{liu2016not}, cannot distinguish between outputs of medium and good quality \cite{novikova2017we}, do not reflect genuine quality improvements in the model output \cite{mathur2020tangled} or nuanced quality distinctions \cite{fomicheva2019taking}, and generally are not a good way to perform evaluation even when good quality references are available as false conclusions can be drawn \cite{vstajner2015deeper}. For task-specific applications, it is reported that word-overlap metrics are more effective for question answering and machine translation, while for dialogue generation and text summarization they present little to no correlation with human judgements \cite{liu2016not}, \cite{novikova2017we}, \cite{kryscinski2019neural}. Interestingly, in machine translation these metrics do better on average at evaluating high quality samples as opposed to low quality samples \cite{fomicheva2019taking}, given that it is difficult to draw meaningful conclusions regarding output quality when there are very few candidate-reference matches. In addition, BLEU is not suitable for the evaluation of text simplification \cite{sulem2018bleu} and document generation \cite{wiseman2017challenges},  and cannot judge the rhythm, meter, creativity, syntactic and semantic coherence in poetry generation \cite{ghazvininejad2017hafez}. Moreover, optimizing discrete metrics such as ROUGE in a reinforcement learning setting does not does not necessarily guarantee an increase in quality, readability and relevance of the generated output \cite{liu2016not}, \cite{paulus2018deep}. Furthermore, BLEU, ROUGE and METEOR are inversely correlated with diversity \cite{sultan2020importance}.

Nevertheless, there are also studies which report high system level correlations for these metrics with human judgements \cite{sulem2018semantic}, \cite{snover2006study}, \cite{anderson2016spice}, at the system-level \cite{reiter2009investigation}, \cite{specia2010machine}, \cite{ma2019results}, sentence-level \cite{fomicheva2019taking} and on worse quality samples \cite{novikova2017we}. Automatic evaluation metrics are more reliable at evaluating the output of neural machine translation models and less reliable at evaluating  conventional statistical translation models, mainly due to differences in translation errors \cite{fomicheva2019taking}. BLEU and METEOR correlate the most with human judgments of grammaticality and meaning preservation, whereas text simplicity is best evaluated by basic length-based metrics \cite{martin2019reference}. ROUGE and its variants is found to agree with manual evaluations of text summarization 
\cite{owczarzak2012assessment}, \cite{rankel2013decade}. In addition, FKBLEU \cite{xu2016optimizing}, iBLEU \cite{sun2012joint} capture the adequacy and
diversity of the generated paraphrase sentence. Metrics assessing inexact matches are also proposed, for eg. TINE \cite{rios2011tine}. 

Estimating the quality of generated text does not require a set of human-written references when it is cast into a prediction task based on features learnt from the training data \cite{specia2010machine}. Reference-less automatic evaluation is also proposed in SAMSA \cite{sulem2018semantic}, which uses semantic parsing on the source side to assess simplification quality. Alternatively, evaluation without requiring references is carried by computing the similarity between the generated output with the source documents in text summarization \cite{louis2013automatically}, \cite{steinberger2012evaluation}.

\paragraph{Grammar-based metrics} The use of grammar-based evaluation metrics has been studied in machine translation \cite{gimenez2008smorgasbord}, grammatical error correction \cite{napoles2016there}, and proposed for the evaluation of generated texts in \cite{novikova2017we}. The authors use the number of mispelings and the Stanford parser score as a crude proxy for the grammaticality of a sentence, in combination with standard readability metrics. Compared to word-overlap metrics, grammar-based metrics do not require a corpus of human-written references, however they fail to establish how relevant the output is to the input.

\paragraph{Perplexity} Perplexity \cite{jelinek1977perplexity} is commonly used to evaluate and compare language models, and measures the average number of words the model is uncertain about when making a prediction. Nevertheless, perplexity is a a model dependent metric, and ``how
likely a sentence is generated by a given model'' is
not directly comparable across different models. Perplexity based evaluation metrics are proposed to measure the fluency and diversity of the generated samples. Reverse Perplexity and Forward Perplexity  \cite{kim2017adversarially} scores are calculated by training language models on synthetic samples, respectively real samples, and measuring the perplexity of the trained model on real samples, respectively generated samples. The Forward Perplexity score captures precision of the generative model, however it is biased in cases when the model repetitively
generates only a few highly likely sentences that yield high scores. The Reverse Perplexity score is dependent upon the quality of the data sample which serves as a proxy for the true data distribution, and the capacity of the language model.

Nevertheless, perplexity is shown to be an inadequate measure of quality \cite{theis2016note}, \cite{fedus2018maskgan}. Likelihoods do not necessarily correspond well to sample quality due to the fact that models with high likelihood can generate low-quality samples, and conversely samples of good quality can present low likelihood. Moreover, infinite perplexity can still be obtained from a perfect model even when its ability to generate test sentences is removed  \cite{hashimoto2019unifying}. Finally, perplexity cannot detect mode collapse in GANs and comparing GAN models based on perplexity puts them at disadvantage with other models since they do not optimize for this objective.

\paragraph{Distance-based metrics} Levenstein distance \cite{levenshtein1966binary}, also known as word edit distance or word error rate \cite{niessen2000evaluation}, quantifies the minimum amount of editing (in terms of additions, deletions and paraphrasing operations) a human would have to perform to convert a hypothesis sentence into its closest reference sentence. TER \cite{snover2006study} normalizes the number of edit operations with the average number of words in the reference, while  TER-Plus \cite{snover2009fluency} relaxes the exact word match assumptions by also counting candidate words that share a stem, are synonyms or paraphrases of the reference words. ITER \cite{panja2018iter} includes stem matching, optimizable
edit costs and improved normalization. PER \cite{tillmann1997accelerated} computes position-independent word error rate at the word level. CDER \cite{leusch2006cder} is combines edit
distance with block reorderings. CharacTER \cite{wang2016character} and EED \cite{stanchev2019eed} extend the edit distance at character
level. Jensen-Shannon divergence compares the underlying probability distributions of n-grams in system summaries
and source documents \cite{lin2006information}, \cite{louis2013automatically}.

Inspired by distance based metrics such as Inception Score (IS) \cite{salimans2016improved} and Fr\'echet Inception Distance (FID) \cite{heusel2017gans} widely used to measure the similarity between real and generated samples in computer vision, Fr\'echet InferSent Distance (FISD) \cite{semeniuta2018accurate} is the equivalent of FID for text evaluation purposes. The FID metric is designed to capture both the quality and diversity of the generated samples by measuring the distance in the embedding space between distributions of features extracted from real and generated samples, nevertheless it does not differentiate the fidelity and diversity aspects of the generated output \cite{naeem2020reliable}. Kernel Inception Distance (KID) \cite{binkowski2018demystifying} is used to measure convergence in GANs through an unbiased estimator independent of sample size. Word Mover's distance (WMD) \cite{kusner2015word} treats documents as bags of embeddings and measures the semantic distance between two texts by computing the amount of flow traveling between embedded words in two documents after aligning semantically  
similar words.
Cosine similarity in the embedding space is used to measure distances between source and target sentences in neural style transfer and quantify the content preservation rate \cite{fu2018style}. RUBER \cite{tao2018ruber} is used for the evaluation of dialogue systems and measures embedding space cosine similarity between a generated response and its query in conversational tasks.

\paragraph{Discriminative Evaluation} Learnt discriminative models are analogous to
learning the discriminator in GANs \cite{goodfellow2014generative}. Based on the two-sample tests \cite{lehmann2006testing} in statistics which summarize differences between two samples into a real-valued test statistic, the goal is to estimate whether two samples $S_p \sim P^n$ and $S_q \sim Q^m$ are drawn from the same data distribution. If $P=Q$, the test accuracy of a binary classifier trained on data samples drawn from the two distributions would remain near-chance level, while if $P \neq Q$ the classifier reveals distributional differences between $S_p$ and $S_q$.  To this end, a classification model is trained with human-written (real) and machine-generated (fake) data samples and its classification accuracy on the test set is used to estimate the quality of the generated samples \cite{bowman2015generating},
\cite{kannan2017adversarial}, \cite{li2017adversarial}, \cite{hodosh2016focused}, 
 \cite{lopez2016revisiting}, \cite{im2018quantitatively}, \cite{ravuri2019classification}. Nevertheless, this approach requires (re-)training a classifier whenever a new generative model is considered and might be biased in cases when the real and fake distributions differ in just one dimension, yielding high overall accuracy but nonetheless assigning lower quality to a superior model \cite{sajjadi2018assessing}.    

Class-conditional GAN architectures are compared by means of evaluation metrics that measure the difference between the learned (generated) and the target (real) distributions. GAN-train and GAN-test \cite{shmelkov2018good} train a classification network on synthetic/real samples generated by a GAN model and evaluate its classification performance on a test set consisting of real-world/generated examples. GAN-train is analogous to recall, while GAN-test is similar to precision. A similar approach is proposed in \cite{ravuri2019classification}, where the Classification Accuracy Score (CAS) measures the performance of a classifier trained on synthetic data at inferring the class labels of real data samples. The metric allows to understand limitations and deficiencies of the generative model. Classification accuracy is also used to measure transfer strength in neural style transfer  \cite{shen2017style}
\cite{fu2018style}, \cite{zhou2018emotional}. 
LEIC \cite{cui2018learning} is used in image captioning to predict if a caption is human-written
or machine-generated.  Furthermore, classification models are also built to distinguish human reference translations from machine translations \cite{corston2001machine}, \cite{kulesza2004learning}, \cite{gamon2005sentence}. 

\paragraph{Precision, Recall and F1 score} are used to measure the distance of the generated samples to the real data manifold \cite{lucic2018gans}. When precision is high, the generated samples are close to the data manifold, and when recall is high, the generator outputs samples that cover the manifold well. Metrics that aggregate precision and recall such as $F_\beta$, a generalization of the $F_1$ score, are used to quantify the relative importance of precision and recall \cite{sajjadi2018assessing}. Nevertheless, the data manifold of non-synthetic data is unknown and therefore impossible to compute in practice.

\paragraph{Readability Metrics} Flesch-Kincaid Grade Level \cite{flesch1948new}, \cite{kincaid1975derivation} and Flesch Reading Ease \cite{flesch1979write} are used to account for simplicity and  measure the reading difficulty of a piece of text. Both metrics are computed as linear combinations of the number of words per sentence and number of syllables per word with different weighting factors. Nevertheless, even though these metrics are frequently used to measure readability, they should not be used on their own but in combination with metrics able to capture the grammaticality and meaning preservation of the generated output \cite{wubben2012sentence}. 

\paragraph{Diversity Metrics} There are many tasks in which it is desirable to generate a set of diverse outputs, such as in story generation to provide multiple continuations for a story prompt \cite{clark2018creative}, in image captioning to capture different perspectives about an image \cite{krause2017hierarchical}, in text reranking algorithms to select best candidate responses and improve user personalization in open-ended dialogue generation and machine translation \cite{li2015diversity}, and in question generation to produce more accurate answers \cite{sultan2020importance}. In the literature diversity of the generated text is regarded from multiple perspectives, on the one hand considering diversity as a measure of how different generated sentences are from each other in terms of word choice, topic and meaning \cite{vijayakumar2016diverse}, \cite{gimpel2013systematic}, \cite{ippolito2018comparison}, and on the other hand accounting for the level of sentence interestingness or unlikeliness \cite{hashimoto2019unifying}. 

Perplexity on a reference set, $n$-gram diversity \cite{li2016diversity} and Self-BLEU \cite{zhu2018texygen} are commonly used measures of the diversity of the generated samples. In addition, Backward-BLEU \cite{shi2018toward} evaluates test data using the generated samples as reference; the higher the score the more diverse the generator output. Lexical diversity \cite{bache2013text} calculates the ratio of unique tokens to the total number of generated tokens. Similarly, Distinct-$k$ or Dist-$k$ \cite{li2016diversity} measures the total number of unique $k$-grams normalized by the total number of generated $k$-gram tokens to avoid favoring long
sentences. Nevertheless, the Dist-$k$ metric ignores the fact that  infrequent $k$-grams contribute more to
diversity than frequent ones and assign same weight to all $k$-grams that appear at least once. Entropy based metrics such as Ent-$k$ \cite{zhang2018generating} are proposed to reflect the frequency difference of $k$-grams and to analyze the information content
of the generated responses in dialogue systems \cite{serban2017hierarchical}, \cite{mou2016sequence}.

\paragraph{Learnt Evaluation Metrics based on Continuous Representations} 
Unlike traditional evaluation metrics based on heuristics, learnable metrics train machine learning models on human annotated datasets to learn a scoring function that reproduces human judgements.
Traditional machine learning models can incorporate human-specified attributes and  handcrafted features, while neural network based approaches work in an end-to-end fashion. In what follows we provide an overview of machine learning based evaluation metrics.

\begin{itemize}
    \item \textit{Fully-learnt metrics} leverage existing datasets of human ratings to learn automated evaluation metrics that fit the human data distribution. In addition, these metrics can be tuned to measure specific properties of the generated texts, such as fluency, style, grammaticality, fidelity, etc. 
    
    MTeRater and MTeRater-Plus \cite{parton2011rating} learn a ranking model for scoring machine translation candidates. A similar ranking approach to evaluating machine translation outputs is adopted in \cite{avramidis2011evaluate}. Machine translation evaluation is approached as a regression task based on linguistic features extracted from the source sentence and its translation \cite{specia2010machine}. 
    BEER \cite{stanojevic2014beer} trains a linear regression model by combining sub-word features (character n-grams) with global word order features (skip bigrams). Linear regression based on human judgements is used to learn a model for scoring system summaries in \cite{peyrard2017learning}. RUSE \cite{shimanaka2018ruse} combines three universal sentence embeddings in a multi-layer perceptron regressor model. ESIM \cite{chen2017enhanced}, \cite{mathur2019putting} feeds the encoded representations of the candidate and the reference sentence into a feedforward regressor. BLEURT \cite{sellam2020bleurt} does quality evaluation by incorporating lexical and semantic pre-training signals and fine-tuning BERT \cite{devlin2018bert} on human ratings datasets for similarity score prediction. MAUDE \cite{sinha2020learning} is proposed for the evaluation of online dialogue conversations and works by leveraging sentence representations from the BERT pre-trained language model to train text encoders which can distinguish between valid dialogue responses and generated negative examples.
    
    Models trained on human judgements are used to predict human scores to dialogue responses. ROSE \cite{conroy2008mind} is a linear combination of ROUGE \cite{lin2004looking} based metrics designed to maximize correlation with human responsiveness. A voting based regression model is proposed to score summaries in  \cite{hirao2007supervised}. Regression based models are also used as a sentence-level metric of machine translation quality \cite{quirk2004training}, \cite{albrecht2007regression}, \cite{albrecht2007re}, \cite{gimenez2008heterogeneous}, \cite{specia2009estimating}. ADEM \cite{lowe2017towards} learns to mimic human judgements in dialogue systems by training a hierarchical RNN encoder to capture the similarity between the dialogue context, the generated model response and human-written reference responses. PARADISE \cite{walker1997paradise} is one of the first learnt evaluation metrics  for the evaluation of task-based dialogue systems.  
    
    \item \textit{Hybrid metrics} combine learnt elements with human-defined logical rules, for example, contextual embeddings with token alignment rules. These metrics are robust to training/ testing data distributing drifts and can work even when limited training data is available. ROUGE \cite{lin2004looking} is enhanced with word embeddings in ROUGE-WE \cite{ng2015better} to capture semantic similarities between words beyond surface lexicographic matches. Human judgements are elicited to extract sets of words with similar meanings for summary evaluation  with the Pyramid scoring scheme \cite{harnly2005automation}, and later extended to fully automated evaluation \cite{yang2016peak}.  YiSi \cite{lo2019yisi} and MEANT \cite{lo2011meant} measure translation quality by matching semantic frames. BERTscore \cite{zhang2019bertscore} evaluates generated text against gold standard references using soft-string similarity matches (i.e. cosine similarity) computed on pre-trained contextualized BERT \cite{devlin2018bert} token embeddings. MoverScore \cite{zhao2019moverscore} combines contextualized representations of system and reference texts with semantic measures of distance  computed using Word Mover’s Distance \cite{kusner2015word}. Furthermore, Word Mover’s Distance is extended to evaluate multi-sentence texts in \cite{clark2019sentence}.  Transformers-based Language Models \cite{kane2019towards} such as RoBERTa \cite{liu2019roberta} are fine-tuned to predict sentence similarity, logical entailment and robustness to grammatical errors for text evaluation purposes. Human and statistical evaluation are combined in HUSE \cite{hashimoto2019unifying}, an evaluation framework which estimates the optimal error rate of predicting whether a piece of text is human-written or machine-generated. Similarly, automatic metrics are combined with human evaluation to infer an unbiased estimator based on control variates which averages differences between human judgments and automatic metrics rather than averaging the human judgments alone \cite{chaganty2018price}.
However, a limitation of such learned evaluation metrics is that they do not generalize well across different systems \cite{chaganty2018price}.

    
\end{itemize}

\subsection{Extrinsic Evaluation}
\label{extrinsic_eval}

Extrinsic evaluation measures the effectiveness of the generated texts on downstream natural language processing tasks or directly on end users. Consequently, extrinsic evaluation is considered the most meaningful type of evaluation in NLG and is generally more useful than intrinsic evaluation \cite{reiter2009investigation}, however extrinsic evaluations are less frequently carried in the literature as they are cost and time intensive, and require careful design.

Extrinsic evaluation methods can be categorized into system-purpose-success and user-type-success metrics \cite{belz2014towards}. 
System-type success metrics quantify the performance of the generated texts on downstream tasks such as information retrieval \cite{fujii2009evaluating}, information extraction \cite{parton2009comparing}, question answering and reading comprehension \cite{jones2007ilr}. User-type success metrics measure the impact of the system on real users as the extent to which it helps them achieve the task it was designed for. Extrinsic evaluations are commonly used in evaluating the performance of task-oriented dialogue agents designed to carry short conversations with human users and assist them in accomplishing a particular goal \cite{deriu2020survey}. 

User performance on a specific task is a direct indicator of text quality \cite{young1999using}, \cite{mani1999tipster}, \cite{di2002diag}, \cite{carenini2006generating}, \cite{hastie2016evaluation}. NLG texts are shown to assist humans in decision-making under uncertainty \cite{gkatzia2016natural}.  Nevertheless, task based evaluations
can be expensive and time-consuming to carry, and results obtained depend on the good will of the participants in the study. Moreover, it is hard to generalize results to new tasks, especially if there is little to no correlation between them. Finally, not every piece of text has a clear function, therefore in some cases a relevant task may not be readily available. 

\subsection{Discussion - NLG Evaluation}
\label{eval_discussion}

In this section we have introduced a wide diversity of metrics for the evaluation of the generated language. As the field of natural language generation is advancing at a fast pace, evaluation becomes critical for measuring progress and conducting fair comparisons between generative models. While many automated evaluation metrics are well established for judging specific natural language tasks, such as BLEU for machine translation, ROUGE and METEOR for text summarization, SARI for text simplification, CIDEr and SPICE for image captioning, there is no universal metric that fits all natural language generation tasks and captures all desirable properties of language. To this end, it is necessary to rely on multiple metrics that reflect different textual attributes such as grammaticality, fluency, coherence, readability, diversity, etc. when conducting language evaluations. However, small changes in the scores reported by these automatic evaluation metrics are not reliable to draw definite conclusions \cite{mathur2020tangled}. Human evaluations remain the gold-standard in natural language generation and automated evaluation metrics can only be used as a proxy for human judgements only when there is reasonable correlation with human decisions. Ideally, automated evaluations are carried simultaneously with human annotation studies, and not as a replacement of human evaluations. 

While progress has been made recently on proposing new evaluation metrics to assess the output of natural language generation systems, more robust evaluation procedures are needed   \cite{novikova2017we}. Moving beyond traditional evaluation metrics that only account for shallow surface form comparisons between the generated texts and gold-standard reference texts, emerging directions in evaluating natural language generation output are focusing on conducting semantic comparisons to achieve better
correlation with human judgments \cite{zhao2019moverscore}. Evaluation metrics based on word and sentence-level embeddings trained from large-scale data and which capture semantic variations show promise in having scalable, cheap, fast and realiable  automated evaluation \cite{ma2019results}. Robust evaluation metrics should also incorporate context \cite{tian2017make}, and account for diversity of content and the presence of rare words which are found to
be more indicative for sentence similarity than common words \cite{zhang2019bertscore}. Evaluating long texts poses special challenges in terms of assessing long-term inter-sentence or inter-paragraph coherence, correctness, fluency, style and semantics, diversity, creativity. It is desirable to have new metrics that are tailored for the evaluation of long texts in particular accounting for these criteria.
In addition, reference-less evaluation of the generated output is an important research direction for tasks such as machine translation, text simplification or dialogue generation when no gold-standard reference data is available \cite{novikova2017we}, \cite{shimanaka2018ruse}. The reference-less quality estimation approach relies on neural networks to predict a
quality score for the generated output by comparing it to the source meaning representation only, therefore presenting the benefit of less  resources invested in collecting expensive human-written annotations.   Moreover, meaningful extrinsic evaluation metrics that measure the contribution of the generated language to task success in a variety of scenarios represent an important future research direction.

Finally, metrics that evaluate the interpretability of neural network models, are able to explain the decisions made (especially valid for metrics based on large pre-trained models) and measure the fairness of generated texts are needed to ensure unbiased, responsible and ethical usage of the natural language generation technology for societal benefit, while combating any of its potential malicious deployments \cite{ippolito2020automatic}, \cite{kreps2020all}. Interpretable explanations can also help determine how much factual knowledge is encoded within the latent parameters of
the model, typically inaccessible to inspection and interpretation, and to what extent this information is memorized from the training corpora \cite{petroni2019language}, \cite{verga2020facts}.  

In parallel with our work, evaluation methods for text generation are also reviewed in \cite{celikyilmaz2020evaluation}, offering a complementary perspective on approaches to natural language evaluation. 







\section{Conclusion}
\label{conclusion}

In the present work we have formally defined the problem of natural language generation at particular contexts and in a variety of natural language processing tasks. In addition, we have presented diverse generative models based on neural networks employed for natural language generation, including recurrent neural networks, sequence-to-sequence models, VAEs, GANs, memory and transfer learning architectures for which we summarized the latest advances focused on language generation. Moreover, we have included a comprehensive overview of methods for evaluating the quality of the generated texts. Given the latest development and the rapid advances in the field, a lot of progress has been made in recent years in both natural language generation and evaluation. Nevertheless, there are still many open challenges to address, including improving generalization to produce novel outputs beyond just memorizing training set examples, generating long-term coherent and diverse texts conditioned or constrained on particular attributes and stylistic properties, learning from few examples in low-resource settings, ensuring fair, ethical and socially responsible uses of the generated text and improving the accountability, explainability and transparency of natural language generative systems.

Evaluation of the generated output is crucial for improving the performance of generative models of natural language, nevertheless it largely remains an open challenge. Human evaluations represent the gold-standard for assessing the quality of machine-generated texts, and automated evaluation metrics should be used with caution only when they present reasonable correlation with human judgements as a complement to human annotations and not as a replacement. Since no automated metric captures all desirable properties of generated text, ideally multiple automated metrics are used simultaneously to capture fine-grained textual attributes such as fluency, readability, coherence, correctness, diversity, etc. Promising directions for developing new evaluation metrics are focused on training neural models to perform reference-less semantic evaluations in the embedding space by means of comparing  the generated output with the source input, as opposed to collecting expensive human-written ground-truth annotations for every task. We also hope to see more focus on task-specific extrinsic evaluation metrics, as well as evaluation metrics which ensure the generated texts are fair, unbiased and do not encode societal stereotypes.


In this survey we have summarized the most recent developments in neural language generation in terms of problem formulation, methods and evaluation. We hope it serves as a useful resource for anyone interested in learning and advancing this fascinating field of research.






\section*{Acknowledgments}

This work was in part supported by the National Science Foundation under grant number 1633370. \\

\bibliography{acl2020}
\bibliographystyle{acl_natbib}
\end{document}